\documentclass[11pt]{article}

\usepackage[final]{acl}

\usepackage[utf8]{inputenc}
\usepackage[T1]{fontenc}
\usepackage{times}
\usepackage{latexsym}
\usepackage{booktabs}
\usepackage{amsfonts}
\usepackage{amsmath}
\usepackage{amssymb}
\usepackage{graphicx}
\usepackage{multirow}
\usepackage{natbib}
\usepackage{float}
\usepackage{tikz}
\usepackage{amsthm}
\newtheorem*{remark}{Remark}
\usetikzlibrary{shapes,arrows.meta,positioning,fit,calc,decorations.pathreplacing}

\title{Riemannian Geometry for Pre-trained Language Model Embeddings}

\author{
{\bf Szczepan Konior$^{1}$\quad Alexandre Quemy$^{2}$\quad Przemys{\l}aw Klocek$^{1}$} \\[1pt]
{\bf Bart{\l}omiej Sobieski$^{3,4}$\quad Gr\'egoire Cattan$^{1}$} \\[2pt]
$^{1}$IBM Automation and AI, Krakow, Poland \quad $^{2}$Hother, Krakow, Poland \\[2pt]
$^{3}$University of Warsaw \quad $^{4}$Centre for Credible AI, Warsaw University of Technology \\[2pt]
{\small\texttt{\{name.surname\}@ibm.com}} \\
{\small\texttt{alexandre@hother.io}} \\
{\small\texttt{sobieski.bartlomiej.jan@gmail.com}}
}

\begin{document}
\maketitle

\begin{abstract}
Understanding the geometric structure of pre-trained language model embeddings matters for interpretability and safety. We ask whether sentence-level classification signal lives in the Riemannian geometry of contextual token embeddings, and probe it by extracting per-token pullback metrics from a learned encoder's analytical Jacobian and aggregating them with the Fr\'echet mean on the symmetric positive definite (SPD) manifold, we call this procedure Riemannian Mean Pooling (RMP). Across three datasets with non-trivial linguistic structure (CoLA, CREAK, RTE), RMP outperforms Euclidean mean pooling, while on FEVER-Symmetric, a benchmark constructed to remove annotation-driven lexical artifacts, the method correctly stays at chance. Ablations show that a randomly initialised encoder combined with Fr\'echet aggregation already beats Euclidean pooling before any training, localising the source of the gain to the geometric aggregation rather than to learned manifold structure. The trained encoder adds a further 1.7 to 2.2 AUC points, consistently across the signal-bearing datasets rather than on any one of them.
\end{abstract}

\section{Introduction}

Pre-trained language models achieve remarkable success in natural language understanding, yet the geometric structure of their internal representations remains an open challenge. Most analyses treat token embeddings as points in flat Euclidean space, but three lines of evidence motivate departing from that assumption. First, hierarchical structure in language admits lower-distortion embeddings in negatively curved spaces~\citep{nickel2017poincare,nickel2018lorentz,sala2018representation,park2025hierarchical} and is reflected in non-standard inner products on the embedding space~\citep{park2024linear,marks2024geometry}. Second, contextualized representations in BERT, ELMo, and GPT-2 are strongly anisotropic, occupying narrow cones not invariant under Euclidean transformation~\citep{ethayarajh2019contextual,lee2025intrinsic}. Third, token embedding spaces may also fail to form smooth manifolds globally, even if local-manifold-like structure still supports geometric analysis at the neighbourhood level~\citep{robinson2025manifoldhypothesis}.

This work investigates whether sentence-level classification signal lives in the Riemannian geometry of token embeddings. We address three questions: (1) Do embeddings exhibit local geometric structure accessible via pullback metrics? (2) Does this geometric structure carry signal beyond what flat Euclidean aggregation extracts? (3) Which component of a geometric aggregation pipeline is responsible for any observed gain: the encoder, the metric, or the aggregation?

We evaluate our approach, \emph{Riemannian Mean Pooling} (RMP), on four diverse NLP tasks: fact verification (FEVER-Symmetric), textual entailment (RTE), grammatical acceptability (CoLA), and commonsense reasoning (CREAK). FEVER-Symmetric is included specifically as a negative control, since it was constructed to remove lexical and annotation artifacts~\citep{FEVERSymmetric}, and we expect any method relying solely on claim-surface geometry to perform at chance.

Any differentiable encoder yields a pullback metric through its Jacobian, but the metric's geometric interpretability depends on the training objective. We use Intrinsic Green's Learning (IGL)~\citep{Quemy2026}\footnote{Reference implementation: \url{https://github.com/hotherio/intrinsic-green-learning}.}, where a closed-form kernel readout replaces the learned decoder, constraining the encoder to coordinates aligned with the smooth structure of the data manifold rather than with arbitrary nonlinear features. Our ablations (Section~\ref{sec:ablations}) test this choice by comparing IGL against random and untrained encoders.

\paragraph{Contributions}
\begin{itemize}
    \item We instantiate a pipeline that extracts per-token pullback metrics from a learned encoder's Jacobian, aggregates them on the SPD manifold via the Fr\'echet mean, and classifies in the tangent space at the population Riemannian mean.
    \item Across three signal-bearing datasets (RTE, CoLA, CREAK), Riemannian aggregation consistently outperforms Euclidean mean pooling under a controlled comparison on identical embeddings. On FEVER-Symmetric, included as a negative control, the method correctly stays at chance.
    \item A randomly initialised encoder combined with Fr\'echet aggregation already beats Euclidean pooling, localising the source of the geometric signal to the aggregation rather than to learned manifold structure. The trained encoder contributes a further 1.7 to 2.2 AUC points on every signal-bearing dataset.
\end{itemize}

\section{Related work}
\label{sec:related}

\paragraph{Riemannian Methods in Neuroscience}

Riemannian geometry is a ubiquitous machine learning technique in brain-computer interfaces (BCIs) and signal processing \citep{congedo_riemannian_2017}. \citet{barachant_multiclass_2012} introduced the use of symmetric positive definite (SPD) matrices as a descriptor for electroencephalography time series, demonstrating that geometric methods can capture the intrinsic structure of signals better than Euclidean approaches. The riemannian minimum-distance-to-means (MDM) and tangent space (TS) classifiers have become ubiquitous tools for neurosciences application, still unbeaten even by the most recent deep learning algorithms \citep{chevallier_largest_2024}, while remaining simple, deterministic, robust to noise, computationally efficient, and prone to transfer learning \citep{andreev_riemannian_2025}.
Beyond the BCI setting, SPD-manifold deep learning has matured: differentiable Fr\'echet-mean layers~\citep{lou2020frechet}, Riemannian batch normalisation for SPD networks~\citep{brooks2019riemannian}, and Riemannian extensions of logistic regression~\citep{chen2024rmlr} now provide standard components for end-to-end SPD pipelines, with recent surveys covering the area~\citep{nguyen2024spd}.

\paragraph{Geometric Approaches in NLP}

In modern NLP architectures, tokens are represented as vectors in a high-dimensional latent space. As tokens propagate through the network layers, their representations evolve inside the Residual Stream, the model's dynamic internal state. This evolution is not purely linear or syntactic, but can be described as a structured trajectory whose geometry changes depending on the semantic context~\citep{manson2025curvedinferenceconcernsensitivegeometry}. Concrete instances of this structure have been demonstrated empirically: linear directions in LLM representations encode space, time, and other conceptual attributes~\citep{gurnee2024language}, and architectural variants explicitly equip attention with non-Euclidean geometry, either via mixed-curvature transformers~\citep{cho2023curve} or via parallel-transport-based attention on manifolds~\citep{liu2025riemannformer}. Recent work has also proposed that the latent space itself can be modelled as a curved spacetime whose curvature reflects linguistic structure~\citep{curvedspacetime2025}.

Current approaches estimate local tangent spaces from k-nearest-neighbor (kNN) patches and derive curvature information from variations between neighboring tangent directions (e.g., \citealp{gao_dynamical_2011,li_curvature-aware_2017}). However, these methods often suffer from numerical instability and sensitivity to sampling density, with metric estimates degrading as the local sample size shrinks~\citep{brand2003charting,SINGER2006128}. A different style of approach first projects the data to a lower-dimensional space (for example via UMAP~\citep{mcinnes_umap_2018}) and computes metrics in the projection, but UMAP preserves local topology rather than metric structure: distances after UMAP are unsuitable as approximations of distances on the original manifold, so subsequent geometric computations operate on a non-isometric image of the data. Recent work has also questioned whether token embeddings form a single connected manifold at all~\citep{robinson2025manifoldhypothesis}. The question we ask here is whether aggregated token geometry nevertheless carries class signal at the sentence level, regardless of whether global manifold structure holds.

\paragraph{Intrinsic Dimension and Manifold Learning}

\citet{sobieski2026localintrinsicdimensionunveils} showed that local intrinsic dimension (LID) reveals hallucinations in diffusion models by identifying instabilities on model-induced manifolds. Their work treats hallucinations as geometric instabilities, proposing Intrinsic Quenching to correct these anomalies. Similarly, \citet{damirchi2026} introduced "Truth as a Trajectory," analysing LLM reasoning through layer-wise geometric metrics rather than static activations. By examining how representations evolve across layers, they uncover geometric invariants distinguishing valid reasoning from spurious behaviour. Beyond these layer-wise trajectory analyses, a body of recent work uses intrinsic dimension as a probe of how LLMs process information across layers: representations expand in early layers and compress in later layers~\citep{valeriani2023geometry}, with characteristic per-layer profiles that include a high-dimensional abstraction phase in the middle layers~\citep{cheng2025emergence}. ID has also been used to compare learning paradigms such as supervised fine-tuning and in-context learning~\citep{janapati-ji-2025-comparative}. Our work complements these perspectives.

\paragraph{Sentence Representation Learning}

Standard sentence-representation approaches such as mean pooling of token embeddings and the BERT \texttt{[CLS]} token implicitly assume flat Euclidean geometry. We replace arithmetic averaging with an SPD-manifold Fr\'echet aggregation. 
Recent work has also highlighted the entanglement of pooled sentence representations~\citep{mechdecomp2025}, motivating geometric alternatives.

\section{Material \& Method}
\label{sec:methodology}

\subsection{Datasets}

We evaluate on four binary-classification benchmarks. \textbf{FEVER-Symmetric}~\citep{FEVERSymmetric} contains 316 fact-verification claims, debiased by Schuster et al.\ against the annotation artifacts of the original FEVER~\citep{Thorne18Fever} by pairing each claim with its negation against shared evidence. We include it as a \emph{negative control}, since methods relying solely on claim-surface features cannot exceed chance by construction. \textbf{RTE}~\citep{wang-etal-2018-glue}: 2{,}490 premise--hypothesis pairs for textual entailment. \textbf{CoLA}~\citep{warstadt2019neural}: 3{,}000 sentences labelled for grammatical acceptability. \textbf{CREAK}~\citep{onoe2021creak}: 3{,}000 commonsense statements requiring world knowledge to verify.

\subsection{Baseline Method: Linear Probe} \label{sec:baseline}

We use \textbf{Linear Probe}~\citep{alain2018understandingintermediatelayersusing} as the headline baseline, referring to this baseline as the Linear Probe throughout and reserving the term \emph{Euclidean mean pooling} for the underlying aggregation operation when contrasting it with the Riemannian alternative. The baseline operates entirely on internal states of the pre-trained model, isolating the contribution of geometric aggregation from any additional trainable parameters or external information sources.

\paragraph{Token embeddings and pooling.} We extract token-level embeddings from the ninth hidden layer of \texttt{BERT-base-uncased}~\citep{bert_2019}, yielding 768-dimensional representations per token. The layer choice is motivated by prior work showing that upper layers encode more abstract semantic features while lower layers capture syntactic information~\citep{rogers_2020}. For single-sentence datasets (FEVER, CoLA, CREAK) we encode the input directly. For the textual-entailment task (RTE) we concatenate premise and hypothesis with separator tokens (\texttt{[CLS] [premise] [SEP] [hypothesis] [SEP]}). Token embeddings are aggregated via arithmetic mean pooling:
\begin{equation}
\mathbf{v}_{\text{sentence}} = \frac{1}{n} \sum_{i=1}^{n} \mathbf{v}_i \in \mathbb{R}^{768},
\end{equation}
with $\{\mathbf{v}_1, \ldots, \mathbf{v}_n\}$ the token embeddings for a sentence with $n$ tokens.

\paragraph{Classification.} Pooled sentence representations are normalised and classified with a logistic regression using scikit-learn's~\citep{scikit-learn} default hyperparameters (no regularisation, LBFGS solver, max-iter 1000). The same classification head is used for the Riemannian pipeline downstream of the Fr\'echet aggregation (Section~\ref{sec:aggregation}), so performance differences attribute cleanly to the aggregation step.

\subsection{Intrinsic Green's Learning}
\label{sec:igl}

\paragraph{Inverse-PDE objective.} IGL~\citep{Quemy2026} is a framework for learning low-dimensional manifold representations through an inverse-PDE formulation. A small MLP encoder $\Psi_\theta : \mathbb{R}^D \to \mathbb{R}^{d_{\max}}$ maps each input $\mathbf{v} \in \mathbb{R}^D$ to a latent code $\boldsymbol{\xi} = \Psi_\theta(\mathbf{v})$. In the encoder's coordinates, the target function (here, the input itself) is expressed as the solution of a linear PDE
\begin{equation}
Lu = s,
\label{eq:igl-pde}
\end{equation}
where $L$ is a fixed operator (a Laplacian in our setup whose Green's function is approximated by a Gaussian kernel) and $s$ is a learned source term.

\paragraph{Closed-form kernel readout.} The PDE is solved in closed form via a Green's-function kernel. In practice, at any latent point $\boldsymbol{\xi}$, the reconstruction is a linear combination of kernel evaluations at $K$ learned anchor points $\{\boldsymbol{\zeta}_k\}_{k=1}^K$,
\begin{equation}
\hat{u}(\boldsymbol{\xi}) = \sum_{k=1}^{K} w_k\, G(\boldsymbol{\xi}, \boldsymbol{\zeta}_k),
\label{eq:igl-readout}
\end{equation}
where $G$ is the Gaussian kernel. The weights $\mathbf{w}$ are not learnable parameters. At every minibatch they are recomputed by ridge regression on the reconstruction targets (variable projection).

\paragraph{Encoder-only optimisation.} IGL has no symmetric decoder. Because $\mathbf{w}$ is closed-form, gradient descent operates only on the encoder parameters $\theta$, and the encoder must find coordinates in which the input admits a low-rank kernel decomposition. Those coordinates align with the smooth structure of the data manifold rather than with arbitrary nonlinear features a learned decoder could compensate for. As a byproduct, because $\Psi_\theta$ is a differentiable MLP, its Jacobian $J(\mathbf{v}) = \partial \Psi_\theta / \partial \mathbf{v}$ is analytically computable, giving direct access to the pullback metric at every input point.

Under the configuration used in this work (MLP encoder, $d_{\max}{=}64$, $K{=}128$ anchors, 4 Gaussian scales per factor), training converges to stable reconstruction. Full architectural details are deferred to Appendix~\ref{app:igl-details}. The full pipeline (training top, inference bottom) is shown schematically in Figure~\ref{fig:token-igl-diagram}. Representative validation-loss curves are shown in Appendix~\ref{app:recon-curve}.

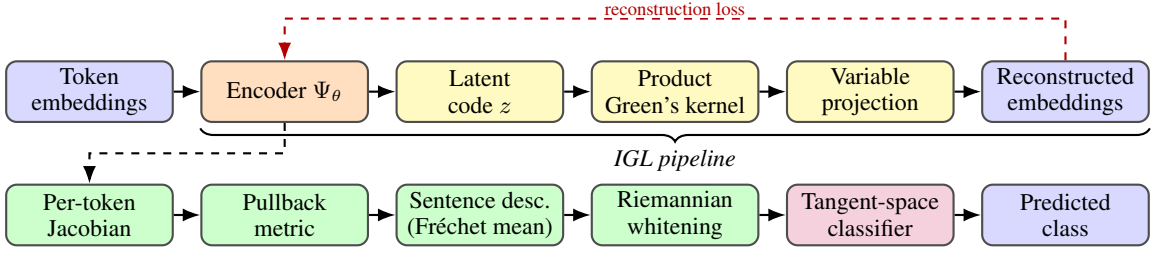
\begin{figure*}[t]
  \centering
  \begin{tikzpicture}[
      every node/.style={font=\small},
      box/.style={rectangle, rounded corners, draw=black!70, thick,
                  minimum width=2.2cm, minimum height=0.8cm, align=center, inner sep=1pt},
      enc/.style={box, fill=orange!25},
      igl/.style={box, fill=yellow!30},
      met/.style={box, fill=green!20},
      cls/.style={box, fill=purple!18},
      io/.style={box, fill=blue!15},
      arrow/.style={-Latex, thick},
      backarrow/.style={-Latex, thick, dashed, red!70!black}
    ]

    \node[io] (tokens) {Token\\embeddings};
    \node[enc, right=0.35cm of tokens] (encoder) {Encoder $\Psi_\theta$};
    \node[igl, right=0.35cm of encoder] (latent) {Latent\\code $z$};
    \node[igl, right=0.35cm of latent] (kernel) {Product\\Green's kernel};
    \node[igl, right=0.35cm of kernel] (vp) {Variable\\projection};
    \node[io, right=0.35cm of vp] (recon) {Reconstructed\\embeddings};

    \draw[arrow] (tokens) -- (encoder);
    \draw[arrow] (encoder) -- (latent);
    \draw[arrow] (latent) -- (kernel);
    \draw[arrow] (kernel) -- (vp);
    \draw[arrow] (vp) -- (recon);

    \draw[backarrow] (recon.north) -- ++(0,0.55) -| (encoder.north)
      node[pos=0.25, above, fill=white, inner sep=1pt,
           font=\scriptsize, text=red!70!black] {reconstruction loss};

    \draw [decorate, decoration={brace, amplitude=5pt, mirror, raise=2pt}, thick]
      (encoder.south west) -- (recon.south east)
      node[midway, yshift=-14pt, font=\small\itshape] {IGL pipeline};

    \node[met, below=0.8cm of tokens] (jac) {Per-token\\Jacobian};
    \node[met, right=0.35cm of jac] (metric) {Pullback\\metric};
    \node[met, right=0.35cm of metric] (frechet) {Sentence desc.\\(Fr\'echet mean)};
    \node[met, right=0.35cm of frechet] (whiten) {Riemannian\\whitening};
    \node[cls, right=0.35cm of whiten] (tsc) {Tangent-space\\classifier};
    \node[io, right=0.35cm of tsc] (label) {Predicted\\class};

    \draw[arrow, dashed] (encoder.south) -- ++(0,-0.4) -| (jac.north);
    \draw[arrow] (jac) -- (metric);
    \draw[arrow] (metric) -- (frechet);
    \draw[arrow] (frechet) -- (whiten);
    \draw[arrow] (whiten) -- (tsc);
    \draw[arrow] (tsc) -- (label);

  \end{tikzpicture}
  \caption{Overview of the pipeline. \textbf{Top row (training):} the encoder $\Psi_\theta$ is trained to reconstruct token embeddings (early stopping with patience=50, $\delta=2 \times 10^{-5}$), mapping them to a low-dimensional latent code from which a product Green's kernel with variable projection recovers the input. The reconstruction loss flows back to the encoder (dashed arrow). \textbf{Bottom row (inference):} once trained, the encoder's Jacobian induces a pullback metric at each token. The per-token metrics are aggregated on the SPD manifold via the Fr\'echet mean to a sentence-level descriptor, preprocessed with Riemannian whitening, projected to the tangent space, and classified. The classification (inference) path carries no gradients and uses the encoder frozen post-training, and only the reconstruction (training) path is differentiable. The full architecture with formulas is given in Figure~\ref{fig:token-igl-diagram-full} of Appendix~\ref{app:igl-details}.}
  \label{fig:token-igl-diagram}
\end{figure*}

\subsection{Pullback metric extraction}
\label{sec:pullback}

After training the IGL encoder $\Psi_\theta: \mathbb{R}^{768} \to \mathbb{R}^{d_{\max}}$ on token-level BERT embeddings $\mathbf{v} \in \mathbb{R}^{768}$ (Section~\ref{sec:igl}), we extract a local Riemannian metric at each token via the \emph{pullback metric}. The intuition is that $\Psi_\theta$ maps the ambient embedding space to a lower-dimensional latent space. The pullback metric measures distances in the input space as they appear after this mapping, capturing how the encoder locally stretches or compresses different directions of the embedding space.

\paragraph{From Jacobian to metric.} For an input $\mathbf{v} \in \mathbb{R}^{768}$, the Jacobian of the encoder is
\begin{equation}
J(\mathbf{v}) = \partial \Psi_\theta / \partial \mathbf{v} \in \mathbb{R}^{d_{\max} \times 768},
\end{equation}
computed in closed form by automatic differentiation through $\Psi_\theta$. Pulling back the ambient Euclidean metric $I_{768}$ through the encoder yields the co-metric on the latent manifold:
\begin{equation}
g^{-1}(\mathbf{v}) = J(\mathbf{v}) \, J(\mathbf{v})^\top \in \mathbb{R}^{d_{\max} \times d_{\max}}.
\label{eq:cometric}
\end{equation}
By construction $g^{-1}(\mathbf{v})$ is symmetric and positive semi-definite. To obtain a metric tensor that is strictly positive-definite (required for the Riemannian operations in Section~\ref{sec:aggregation}), we add a small regulariser before inversion:
\begin{equation}
g(\mathbf{v}) = \left( g^{-1}(\mathbf{v}) + \epsilon I_{d_{\max}} \right)^{-1}, \quad \epsilon = 10^{-6},
\end{equation}
and a further additive regularisation $\lambda I_{d_{\max}}$ with $\lambda = 10^{-2}$ to guarantee numerical stability across the full range of input tokens. The resulting $g(\mathbf{v}) \in \mathcal{S}_{++}^{d_{\max}}$ is the per-token SPD matrix used as input to the sentence-level aggregation.

\begin{remark}
The expression \textnormal{pullback metric} is an abuse of language, since $d_{\max} < 768$, $\Psi_\theta$ is a submersion, and \eqref{eq:cometric} is not a metric but a co-metric, obtained by pushing the ambient $I_{768}$ forward along $\Psi_{\theta}$. The strict pullback would be the degenerate $J^\top J$ on $\mathbb{R}^{768}$. Restricting $\Psi_\theta$ to an estimated data manifold $M$ of dimension $m < d_{\max}$ makes it an immersion and replaces \eqref{eq:cometric} with $J P_M J^\top$. Our runs show that it does not improve RMP on any dataset, achieving the same or worse performance than the unrestricted metric. We report that in Appendix~\ref{app:immersion}.
\end{remark}

\paragraph{Why pullback rather than direct estimation?}
The two most direct alternatives, that is, neighbour-based metric estimation and projection-then-metric approaches (e.g. UMAP), suffer from sensitivity to local sampling density and non-isometric distortion, respectively (Section~\ref{sec:related}). The pullback through a trained encoder avoids both issues: the metric is computed analytically from $J$ rather than estimated from neighbours, and its position dependence is a smooth function of the original 768-dimensional input rather than of a non-isometric projection of it. The sensitivity of neighbour-based estimation is measurable on these embeddings and has been reported in Appendix~\ref{app:immersion}. Pulling back an ambient metric through an encoder's Jacobian follows~\citet{hauser2017principles} and~\citet{arvanitidis2018latent}. Recent extensions analyse discretisation of continuous inputs~\citep{liang2025emergent} and score-based extraction of data-manifold geometry~\citep{diepeveen2024score}.

\paragraph{Independence from the training loss.} Because the pullback metric depends only on the encoder's Jacobian rather than on its training loss, the geometric pipeline downstream of $g(\mathbf{v})$ is well-defined for any differentiable encoder. We exploit this in Section~\ref{sec:ablations} to isolate the contribution of the encoder from the contribution of the geometric aggregation.

\subsection{Sentence aggregation and classification}
\label{sec:aggregation}

\paragraph{Token-to-Sentence Aggregation}

Given a sentence with $n$ tokens producing per-token SPD matrices $\{g(\mathbf{v}_1), \ldots, g(\mathbf{v}_n)\} \in \mathcal{S}_{++}^{d_{\max}}$ (Section~\ref{sec:pullback}), we aggregate them into a single sentence-level SPD descriptor using the Fr\'echet mean~\citep{fletcher_2004} on the SPD manifold:
\begin{equation}
G_{\text{sentence}} = \underset{G \in \mathcal{S}_{++}^{d_{\max}}}{\arg\min} \sum_{i=1}^{n} d_{\text{Riem}}^2(G, g(\mathbf{v}_i)),
\end{equation}
with $d_{\text{Riem}}$ the affine-invariant Riemannian distance~\citep{frstner_2000}:
\begin{equation}
d_{\text{Riem}}(G_1, G_2) = \left\| \log(G_1^{-1/2} G_2 G_1^{-1/2}) \right\|_F.
\end{equation}
$G_{\text{sentence}}$ is the geometric counterpart of Euclidean mean pooling, a Fr\'echet-mean point on the SPD manifold rather than an arithmetic mean.

\paragraph{Classification}

The sentence-level SPD matrices are classified in three steps. First, we apply Riemannian whitening, which rescales each matrix to unit determinant while preserving its affine-invariant geometric properties, removing scale differences across samples. Second, we project each whitened matrix to the tangent space at the population Riemannian mean of the training set. The tangent space at a point on the SPD manifold is a flat Euclidean space, so standard linear classifiers apply. Third, we fit a logistic regression on the vectorised tangent-space representations. We use pyRiemann's~\citep{pyriemann} \texttt{TSClassifier} with default hyperparameters, which implements the projection-plus-linear-classifier pipeline.

\subsection{Experimental configuration}

All experiments employed 5-fold stratified cross-validation to ensure balanced class distributions across folds, using two seeds. We report the mean and standard deviation across folds to capture performance variability.

Because the classification pipeline reuses established BCI components (e.g., Fr\'echet mean of SPD matrices, Riemannian whitening, and tangent-space classifier) we adopt the evaluation and statistical protocol of the Mother of all BCI Benchmarks (MOABB)~\citep{aristimunha_mother_2023}. Statistical significance was computed independently for each dataset using significance thresholds of $\alpha < 0.05$ and $\alpha < 0.01$. Effect sizes are reported using the standardised mean difference (SMD), and the aggregated meta-effect across datasets is also provided.

\section{Experimental results}
\label{sec:results}

Table~\ref{tab:multiseed-results} reports two baselines, Linear Probe (Euclidean mean pooling) and CLS Token Aggregation, alongside RMP on the three signal-bearing datasets (CoLA, CREAK, RTE) and the FEVER-Symmetric negative control, with means and standard deviations across the multi-seed cross-validation protocol.

\begin{table*}[ht]
\centering
\small
\setlength{\tabcolsep}{4pt}
\begin{tabular}{llccc}
\toprule
Dataset & Method & Accuracy & F1-Score & AUC \\
\midrule
\multirow{3}{*}{\textbf{FEVER}$^\dagger$}
    & Linear Probe & 0.405 $\pm$ 0.046 & 0.426 $\pm$ 0.036 & 0.425 $\pm$ 0.029 \\
    & CLS Token Aggregation & 0.375 $\pm$ 0.051 & 0.398 $\pm$ 0.053 & 0.402 $\pm$ 0.052 \\
    & Riemannian Mean Pooling & 0.494 $\pm$ 0.043 & 0.513 $\pm$ 0.044 & 0.507 $\pm$ 0.039 \\
\midrule
\multirow{3}{*}{\textbf{CoLA}}
    & Linear Probe & 0.675 $\pm$ 0.018 & 0.678 $\pm$ 0.017 & 0.731 $\pm$ 0.021 \\
    & CLS Token Aggregation & 0.661 $\pm$ 0.014 & 0.662 $\pm$ 0.016 & 0.711 $\pm$ 0.012 \\
    & Riemannian Mean Pooling & \textbf{0.711 $\pm$ 0.014} & \textbf{0.723 $\pm$ 0.014} & \textbf{0.782 $\pm$ 0.021} \\
\midrule
\multirow{3}{*}{\textbf{CREAK}}
    & Linear Probe & 0.607 $\pm$ 0.026 & 0.604 $\pm$ 0.026 & 0.646 $\pm$ 0.032 \\
    & CLS Token Aggregation & 0.593 $\pm$ 0.026 & 0.593 $\pm$ 0.027 & 0.631 $\pm$ 0.024 \\
    & Riemannian Mean Pooling & \textbf{0.633 $\pm$ 0.014} & \textbf{0.628 $\pm$ 0.021} & \textbf{0.685 $\pm$ 0.018} \\
\midrule
\multirow{3}{*}{\textbf{RTE}}
    & Linear Probe & 0.559 $\pm$ 0.014 & 0.554 $\pm$ 0.020 & 0.591 $\pm$ 0.025 \\
    & CLS Token Aggregation & 0.573 $\pm$ 0.028 & 0.581 $\pm$ 0.028 & 0.611 $\pm$ 0.029 \\
    & Riemannian Mean Pooling & \textbf{0.594 $\pm$ 0.021} & \textbf{0.597 $\pm$ 0.018} & \textbf{0.632 $\pm$ 0.025} \\
\bottomrule
\end{tabular}
\caption{Multi-seed meta-analysis results (mean $\pm$ std across 10 evaluations). $^\dagger$FEVER is included as a negative control (Section~\ref{sec:methodology}). Chance-level performance is the expected and correct behaviour, so no method is bolded for this dataset.}
\label{tab:multiseed-results}
\end{table*}

\begin{figure}[t]
  \centering
  \includegraphics[width=0.9\columnwidth]{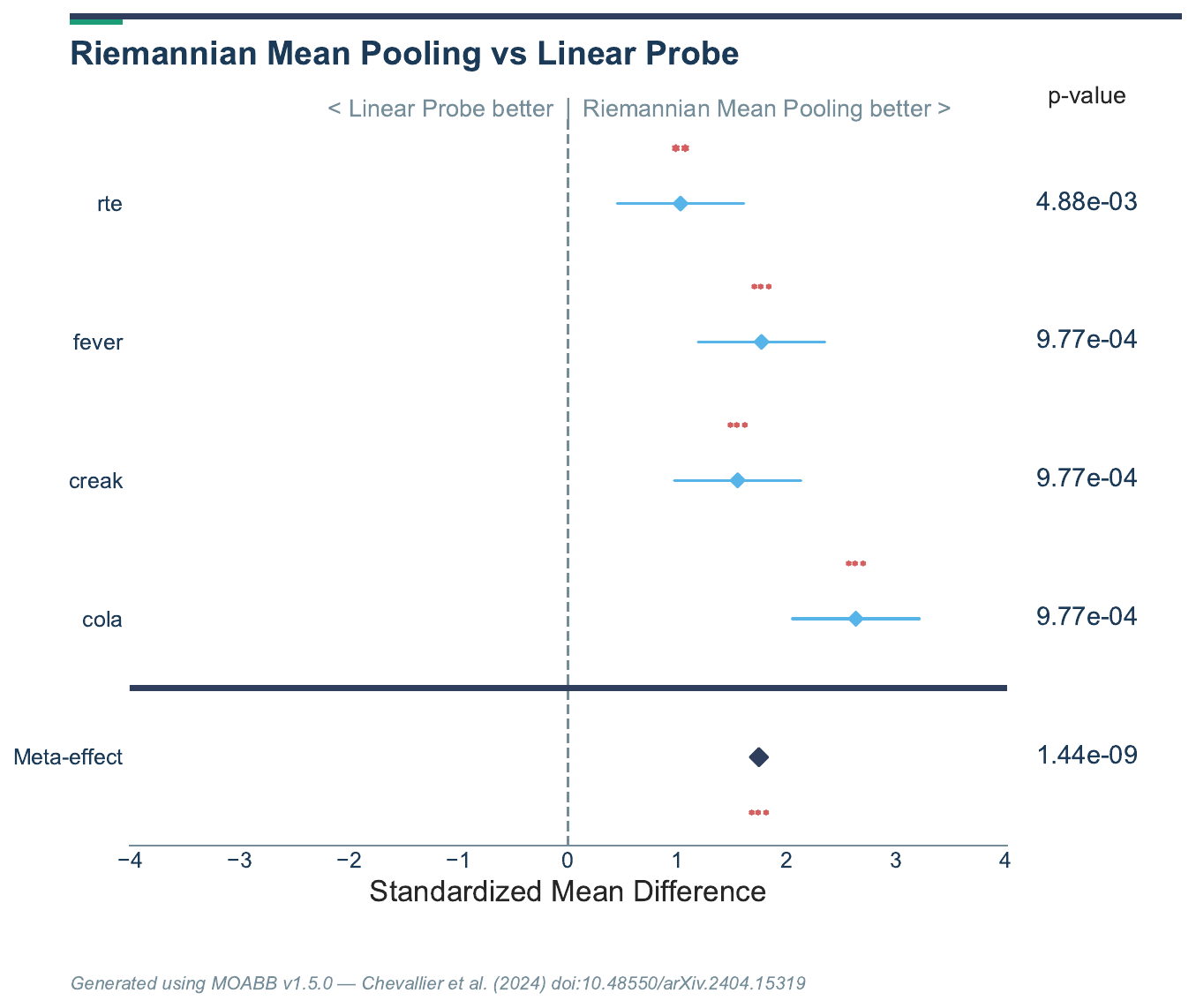}
  \caption{Statistical comparison between Linear Probe and Riemannian Mean Pooling across random seeds. Box plots show the distribution of AUC scores obtained from 10 evaluations (2 seeds $\times$ 5 folds). Corresponding p-values are displayed on the right side of each plot. One (respectively, two and three) red star(s) indicate that the p-value is below the significance threshold of $\alpha < 0.05$ (respectively, $\alpha < 0.01$, $\alpha < 0.001$). FEVER-Symmetric is a negative control (Section~\ref{sec:methodology}), so a chance-level result there is the expected behaviour.}
  \label{fig:multiseed-statistical}
\end{figure}

Figure~\ref{fig:multiseed-statistical} presents forest plot comparing the distribution of AUC scores across all 10 evaluations for each dataset. 

Per-dataset AUC, Accuracy, and F1 bar plots together with the CLS-vs-RMP forest plot (Figures~\ref{fig:multiseed-auc}, \ref{fig:multiseed-accuracy}, \ref{fig:multiseed-f1}, \ref{fig:multiseed-statistical-cls}) are deferred to Appendix~\ref{app:additional-results}. Figure~\ref{fig:multiseed-statistical} and Table~\ref{tab:multiseed-results} carry the headline comparison in the body.

The multi-seed meta-analysis shows RMP outperforming both baselines (Linear Probe and CLS Token Aggregation) on CoLA, CREAK, and RTE, the three datasets with classification signal accessible from claim surface. CLS Token Aggregation underperforms Linear Probe on three of four datasets (FEVER, CoLA, CREAK), with RTE the exception (0.611 vs 0.591 AUC). This replicates the observation of~\citet{reimers2019sentencebert} that vanilla CLS embeddings underperform mean pooling on transfer tasks without task-specific fine-tuning, and provides a second controlled Euclidean baseline against which to measure the Riemannian gain. On FEVER-Symmetric, RMP stays at chance, which is the behaviour expected from a negative control. Its point estimate of $0.507$ AUC nominally exceeds $0.5$, but it falls well inside the 95\,\% interval $[0.419,\,0.579]$ of a label-permutation null over $10{,}000$ draws~\citep{muller-putz_better_2008,maris_nonparametric_2007}, at $p = 0.83$ and a standardised mean difference of $0.16$, below the conventional threshold for a small effect. The two Euclidean baselines do not stay at chance but fall significantly below it. This is further discussed in Section~\ref{sec:discussion}.

\section{Decomposing the geometric gain}
\label{sec:ablations}

\subsection{Ablation Components}

To localise the source of the geometric gain, we ablate the pipeline along two axes: (i) whether the encoder is trained or randomly initialised, and (ii) whether the encoder is the IGL MLP architecture or a simpler random linear projection. \textbf{Encoder-Only} projects tokens through the trained IGL encoder and applies Euclidean mean pooling in latent space (no metric extraction), testing whether the encoder alone matters. \textbf{Frozen Random IGL} replaces the trained encoder with a randomly initialised MLP of identical architecture (256-d hidden, SiLU) and runs the full geometric pipeline, testing whether encoder training matters. \textbf{Random Projection + SiLU} uses a frozen random linear projection $\mathbb{R}^{768} \to \mathbb{R}^{64}$ followed by SiLU, then the full geometric pipeline, testing whether the depth and nonlinearity of the IGL MLP matter even without training.

\subsection{Experimental Results}

The full ablation table (Table~\ref{tab:ablation-results}) and per-comparison forest/bar plots are deferred to Appendix~\ref{app:ablation-figures}, and the qualitative findings are summarised below. In addition, Appendix~\ref{app:concat} reports a further control in which the tangent-space and mean-pooled BERT features are evaluated separately and concatenated under one classifier, and finds that the size of the feature set does not account for the geometric gain.

\subsection{Findings}

Three patterns emerge from Table~\ref{tab:ablation-results}. First, a randomly initialised encoder recovers most of the geometric gain on every signal-bearing dataset (CoLA 0.760 against 0.782, CREAK 0.667 against 0.685, RTE 0.615 against 0.632), indicating that the geometric aggregation, rather than encoder training, carries most of the gain. Second, the trained encoder adds 1.7 to 2.2 AUC points on all three datasets (paired over folds, CoLA gains $+0.022$ at $p = 0.0002$, CREAK $+0.018$ at $p = 0.004$, and RTE $+0.017$ at $p = 0.024$). Third, the simpler Random Projection + SiLU underperforms the random IGL encoder on every signal-bearing dataset, indicating that the depth and nonlinearity of the MLP matter even without training. The FEVER-Symmetric rows are reported for completeness but, by the negative-control framing of Section~\ref{sec:methodology}, are not informative about the source of geometric signal.

\section{Discussion}
\label{sec:discussion}

\subsection{Hypothesis validation}

\paragraph{Geometric advantage and what it tests.} Riemannian aggregation of token-level pullback metrics outperforms Euclidean mean pooling on the three signal-bearing datasets and stays at chance on the FEVER-Symmetric negative control (Table~\ref{tab:multiseed-results} and Figure~\ref{fig:multiseed-statistical}). The gain is largest on CoLA ($\Delta\text{AUC} = 0.051$) and similar on RTE and CREAK ($0.041$ and $0.040$). The ablations localise most of this gain to the geometric aggregation rather than to learned manifold structure. RMP with a random encoder already significantly beats Euclidean pooling on CoLA and CREAK, while the trained encoder adds 1.7 to 2.2 AUC points across all three datasets (Table~\ref{tab:ablation-results}, with per-comparison forest plots in Appendix~\ref{app:ablation-figures}). Rather than concluding that BERT embeddings lie on a globally coherent manifold, the evidence supports \emph{locally manifold-like structure}~\citep{robinson2025manifoldhypothesis}, under which aggregated geometric features yield classification gains even where global manifold structure fails.

\paragraph{Role of the negative control}

The inclusion of FEVER-Symmetric~\citep{FEVERSymmetric} serves a methodological purpose distinct from the other three datasets. Where CoLA, CREAK, and RTE probe whether geometric structure encodes specific linguistic properties, FEVER-Symmetric probes whether the method exploits residual signal in the underlying representations.

RMP stays at chance on it, reaching $0.507$ against a label-permutation null of $0.500$ (10{,}000 permutations, 95\,\% interval $[0.419,\,0.579]$, $p = 0.83$, SMD $0.16$). The two Euclidean baselines perform significantly below chance (Linear Probe reaches $0.425$ and CLS Token Aggregation $0.402$). This phenomenon is explained by our evaluation protocol combined with dataset construction. By dataset construction, we mean that FEVER-Symmetric pairs most of the claims (228 over 316 rows) with a minimal edit carrying the opposite label. In fact, BERT embeddings place a claim and its twin close together in representation space. Therefore, a classifier that operates on claim-surface features will assign both members of a pair the same predicted label. This is an artifact of our evaluation method. Indeed, under five-fold cross-validation a pair is split across training and test four times out of five, so the classifier is trained on one claim and evaluated on its negation 80\% of the time. The member of the pair in the testing set then receives the label of its twin in the training set. This is why both baselines invert on the twinned rows, while RMP staying close to chance is the expected behaviour. Appendix~\ref{app:fever} reports the negative control in full.

\paragraph{What does the pullback metric encode?}

The random-encoder ablations clarify what the pipeline is actually testing. The pullback metric $g(\mathbf{v}) = (JJ^\top + \epsilon I)^{-1}$ captures the local anisotropy of the encoder map, that is, the extent to which a unit input perturbation at $\mathbf{v}$ is stretched or compressed across latent dimensions. For a random MLP with SiLU nonlinearity, $J(\mathbf{v})$ varies smoothly with $\mathbf{v}$ because of the nonlinearity, even though the encoder weights themselves are random. Consequently $g(\mathbf{v})$ encodes the position of $\mathbf{v}$ in input space through the local Jacobian structure of the random transformation, not through any property the encoder learned. Fr\'echet aggregation on the SPD manifold then summarises this local-geometry signal into a sentence-level descriptor. The geometric-aggregation gain over Euclidean pooling thus reflects that token positions in BERT embeddings carry classification signal in their local nonlinear-transformation structure, which Euclidean averaging discards.

\subsection{Limitations and challenges}

\paragraph{Computational cost.} On top of one-time IGL training, each sentence requires a per-token Jacobian and an SPD Fr\'echet-mean iteration, an order of magnitude more arithmetic than Euclidean pooling, though still negligible against LLM inference or gradient-based fine-tuning. Exact wall-clock and FLOP figures are reserved for a scaling study.

\paragraph{Hyperparameter and architectural scope.} Our encoder-width and anchor-count sweep was conducted only on FEVER-Symmetric, so sensitivity on the signal-bearing datasets and to the metric regulariser $\lambda$ is unverified. The encoder objective itself was held fixed at IGL. Alternative objectives that also yield a Jacobian (vanilla autoencoders, VAEs, normalising flows, contrastive encoders) were not swept, although the random-encoder ablation already suggests the specific objective is largely transparent on most tasks.

\paragraph{Scope of the controlled comparison.} The pipeline was evaluated against Euclidean and CLS-token baselines on a single encoder (BERT-base) at a single layer (layer 9) and a fixed projection dimensionality. This is a deliberate methodological choice. The comparison isolates the contribution of geometric aggregation on identical embeddings, which is why fine-tuned BERT and Sentence-BERT~\citep{reimers2019sentencebert} are excluded rather than missing. Absolute classification performance is not the question this paper investigates. The question is whether geometric aggregation extracts signal that flat-Euclidean pooling discards.

\subsection{Future work}

\paragraph{Architecture and layer coverage.} Extending the evaluation to larger and structurally different encoders (RoBERTa, DistilBERT, decoder-only LLMs) and across layers would clarify whether the geometric-aggregation gain is general or specific to BERT-base layer 9, and whether the layer-wise intrinsic-dimension trajectory~\citep{valeriani2023geometry,cheng2025emergence} interacts with where the pullback metric becomes informative.


\paragraph{Richer SPD descriptors.} The Fr\'echet mean is the simplest aggregator on the SPD manifold. Weighted Fr\'echet means, trajectory-based descriptors in the spirit of layer-wise geometric probes~\citep{damirchi2026,mir2025geometry}, and explicit curvature features are natural next steps, with per-token pullback eigenvalues offering a route to relate the geometric signal to interpretable linguistic phenomena.

\section{Conclusion}
\label{sec:conclusion}
The SPD Fr\'echet mean of per-token pullback metrics beats Euclidean mean pooling on CoLA, CREAK and RTE under a controlled comparison on identical embeddings. On FEVER-Symmetric, included as a negative control, it correctly stays at chance. Random-encoder ablations localise most of the gain to the geometric aggregation rather than to learned manifold structure, with the trained encoder contributing a further 1.7 to 2.2 AUC points on every signal-bearing dataset. Together these results support the framing of embeddings as locally manifold-like rather than globally coherent~\citep{robinson2025manifoldhypothesis}, under which aggregated geometric features yield classification gains even where global manifold structure may fail.

\bibliography{biblio}

\twocolumn[{%
  \centering
  \includegraphics[width=\textwidth]{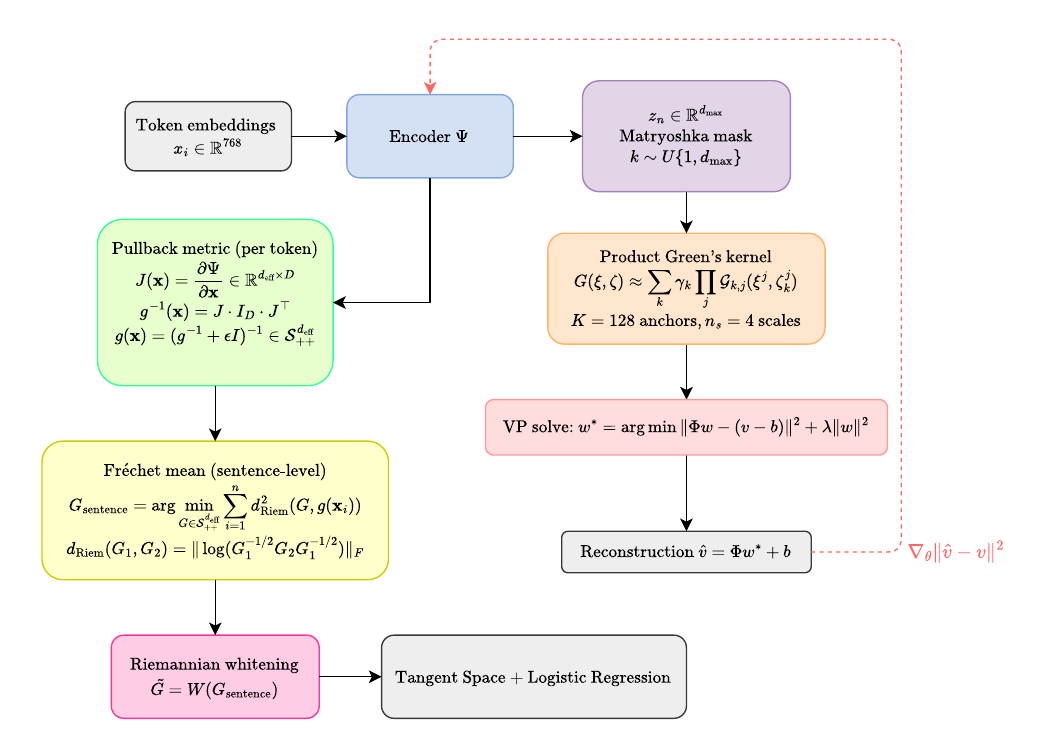}
  \captionof{figure}{Detailed architecture of token-level IGL for sentence classification via RMP. \textbf{Right path (reconstruction):} Token embeddings $\mathbf{v}_i \in \mathbb{R}^{768}$ are encoded to latent space $\mathbf{z} \in \mathbb{R}^{d_{\max}}$ via encoder $\Psi_\theta$, then mapped through Product Green's kernel $G(\xi, \zeta)$ to design matrix $\Phi$. Variable projection (VP) solves for optimal weights $w^* = \arg\min_w \|\Phi w - (y - b)\|^2 + \lambda \|w\|^2$ to reconstruct input $\hat{\mathbf{v}} = \Phi w^* + b$. \textbf{Left path (classification):} The trained encoder's Jacobian $J = \frac{\partial \Psi_\theta}{\partial \mathbf{v}}$ induces pullback metrics $g(\mathbf{v}_i) \in \mathcal{S}_{++}^{d_{\text{max}}}$ at each token. These are aggregated via Fr\'echet mean to sentence-level SPD matrix $G_{\text{sentence}}$, preprocessed with Riemannian whitening, projected to tangent space, vectorised, and classified with logistic regression. \textbf{Solid arrows:} forward flow. \textbf{Dashed red arrow:} reconstruction-loss gradient $\nabla_\theta \|\hat{\mathbf{v}} - \mathbf{v}\|^2$ flowing back to encoder $\Psi_\theta$. The encoder is trained jointly with VP solve until reconstruction converges (early stopping with patience=50, $\delta=2 \times 10^{-5}$), then frozen for metric extraction and classification.}
  \label{fig:token-igl-diagram-full}
  \vspace{\dbltextfloatsep}
}]
\appendix

\section{IGL architecture details}
\label{app:igl-details}

This appendix collects the architectural details of the IGL configuration used in Section~\ref{sec:igl}.

\begin{table*}[!t]
\centering
\small
\begin{tabular}{llcc}
\toprule
Dataset & Method & Accuracy & AUC \\
\midrule
\multirow{6}{*}{\textbf{FEVER}$^\dagger$}
    & Linear Probe & 0.405 $\pm$ 0.046 & 0.425 $\pm$ 0.029 \\
    & CLS Token Aggregation & 0.375 $\pm$ 0.051 & 0.402 $\pm$ 0.052 \\
    & Riemannian Mean Pooling & 0.494 $\pm$ 0.043 & 0.507 $\pm$ 0.039 \\
    & Encoder-Only & 0.506 $\pm$ 0.043 & 0.537 $\pm$ 0.036 \\
    & Frozen Random IGL & 0.426 $\pm$ 0.040 & 0.441 $\pm$ 0.035 \\
    & Random Projection + SiLU & 0.498 $\pm$ 0.080 & 0.515 $\pm$ 0.081 \\
\midrule
\multirow{6}{*}{\textbf{CoLA}}
    & Linear Probe & 0.675 $\pm$ 0.018 & 0.731 $\pm$ 0.021 \\
    & CLS Token Aggregation & 0.661 $\pm$ 0.014 & 0.711 $\pm$ 0.012 \\
    & Riemannian Mean Pooling & \textbf{0.711 $\pm$ 0.014} & \textbf{0.782 $\pm$ 0.021} \\
    & Encoder-Only & 0.686 $\pm$ 0.031 & 0.754 $\pm$ 0.032 \\
    & Frozen Random IGL & 0.695 $\pm$ 0.022 & 0.760 $\pm$ 0.021 \\
    & Random Projection + SiLU & 0.641 $\pm$ 0.024 & 0.692 $\pm$ 0.028 \\
\midrule
\multirow{6}{*}{\textbf{CREAK}}
    & Linear Probe & 0.607 $\pm$ 0.026 & 0.646 $\pm$ 0.032 \\
    & CLS Token Aggregation & 0.593 $\pm$ 0.026 & 0.631 $\pm$ 0.024 \\
    & Riemannian Mean Pooling & \textbf{0.633 $\pm$ 0.014} & \textbf{0.685 $\pm$ 0.018} \\
    & Encoder-Only & 0.623 $\pm$ 0.011 & 0.664 $\pm$ 0.017 \\
    & Frozen Random IGL & 0.623 $\pm$ 0.017 & 0.667 $\pm$ 0.017 \\
    & Random Projection + SiLU & 0.584 $\pm$ 0.023 & 0.621 $\pm$ 0.028 \\
\midrule
\multirow{6}{*}{\textbf{RTE}}
    & Linear Probe & 0.559 $\pm$ 0.014 & 0.591 $\pm$ 0.025 \\
    & CLS Token Aggregation & 0.573 $\pm$ 0.028 & 0.611 $\pm$ 0.029 \\
    & Riemannian Mean Pooling & \textbf{0.594 $\pm$ 0.021} & \textbf{0.632 $\pm$ 0.025} \\
    & Encoder-Only & 0.584 $\pm$ 0.026 & 0.630 $\pm$ 0.031 \\
    & Frozen Random IGL & 0.576 $\pm$ 0.022 & 0.615 $\pm$ 0.025 \\
    & Random Projection + SiLU & 0.561 $\pm$ 0.024 & 0.583 $\pm$ 0.030 \\
\bottomrule
\end{tabular}
\caption{Ablation study results (mean $\pm$ std across 10 evaluations). Bold indicates best per dataset. F1 is omitted from this table (tracks Accuracy closely on binary tasks). The full Accuracy/F1/AUC table is in Table~\ref{tab:multiseed-results}, which is drawn from the same run, so the rows the two tables share are identical by construction. $^\dagger$FEVER is included as a negative control (Section~\ref{sec:methodology}). Rows are reported for completeness but are not informative about the source of geometric signal, so no method is bolded.}
\label{tab:ablation-results}
\end{table*}

\paragraph{Encoder and Matryoshka Truncation}

A small MLP encoder $\Psi_\theta: \mathbb{R}^{D} \rightarrow \mathbb{R}^{d_{\max}}$ maps input to latent codes $\xi_n = \Psi_\theta(v_n)$ with $d_{\max} = 64$. During training, a random truncation $k \sim \text{Uniform}\{1, \ldots, d_{\max}\}$ is sampled and trailing coordinates zeroed out, following Matryoshka representation learning~\citep{kusupati2024matryoshkarepresentationlearning}. This ensures earlier dimensions capture the most informative features, with effective dimension $d_{\text{max}}$ determined post-training from the dimension-loss curve.

\paragraph{Product Green's Kernel}

The reconstruction target is the input $v_n$ itself. We model the
reconstruction by a tensor-factorised product Green's kernel:

\begin{multline}
G(\xi, \zeta) \approx \sum_{k=1}^{K} \gamma_k \prod_{j=1}^{d_{\max}} G_{k,j}(\xi_j, \zeta_j), \\ K = n_{\text{anchors}} = 128 \hspace{2cm}
\end{multline}

Each factor $G_{k,j}$ is a one-dimensional kernel evaluated at the latent coordinate $\xi^j$ and an anchor $\zeta^j_k$, with K anchors placed adaptively. We use Gaussian kernel factors. Each factor is itself a convex mixture of $n_{\text{scales}} = 4$ Gaussian basis functions at learnable bandwidths $\sigma_{s, j}$,

\begin{multline}
G_{k,j}(\xi_j, \zeta_j^k) = \sum_{s=1}^{4} \alpha_s \exp\left(-\frac{(\xi_j - \zeta_j^k)^2}{2\sigma_{s,j}^2}\right), \\
a = \text{softmax}(\cdot) \hspace{2cm}
\end{multline}

\paragraph{Variable-Projection Reconstruction}

Each minibatch solves bias-centered ridge regression for weights, with $\lambda = 10^{-3}$:
\begin{equation}
w^* = \arg\min_w \|\Phi w - (v - b)\|^2 + \lambda \|w\|^2
\end{equation}

The design matrix $\Phi$ costs $O(NKd)$ to compute, while the closed-form solve is $O(NK^2 + K^3)$ and independent of $d$. Weights $w^*$ are recomputed per minibatch (not stored), so gradient descent operates only on encoder parameters $\theta$.

\section{Ablation figures}
\label{app:ablation-figures}

The full ablation-results table and per-comparison forest plots and bar plots for the three ablation contrasts (Section~\ref{sec:ablations}) are collected here for space.

\begin{figure}[H]
  \centering
  \includegraphics[width=\columnwidth]{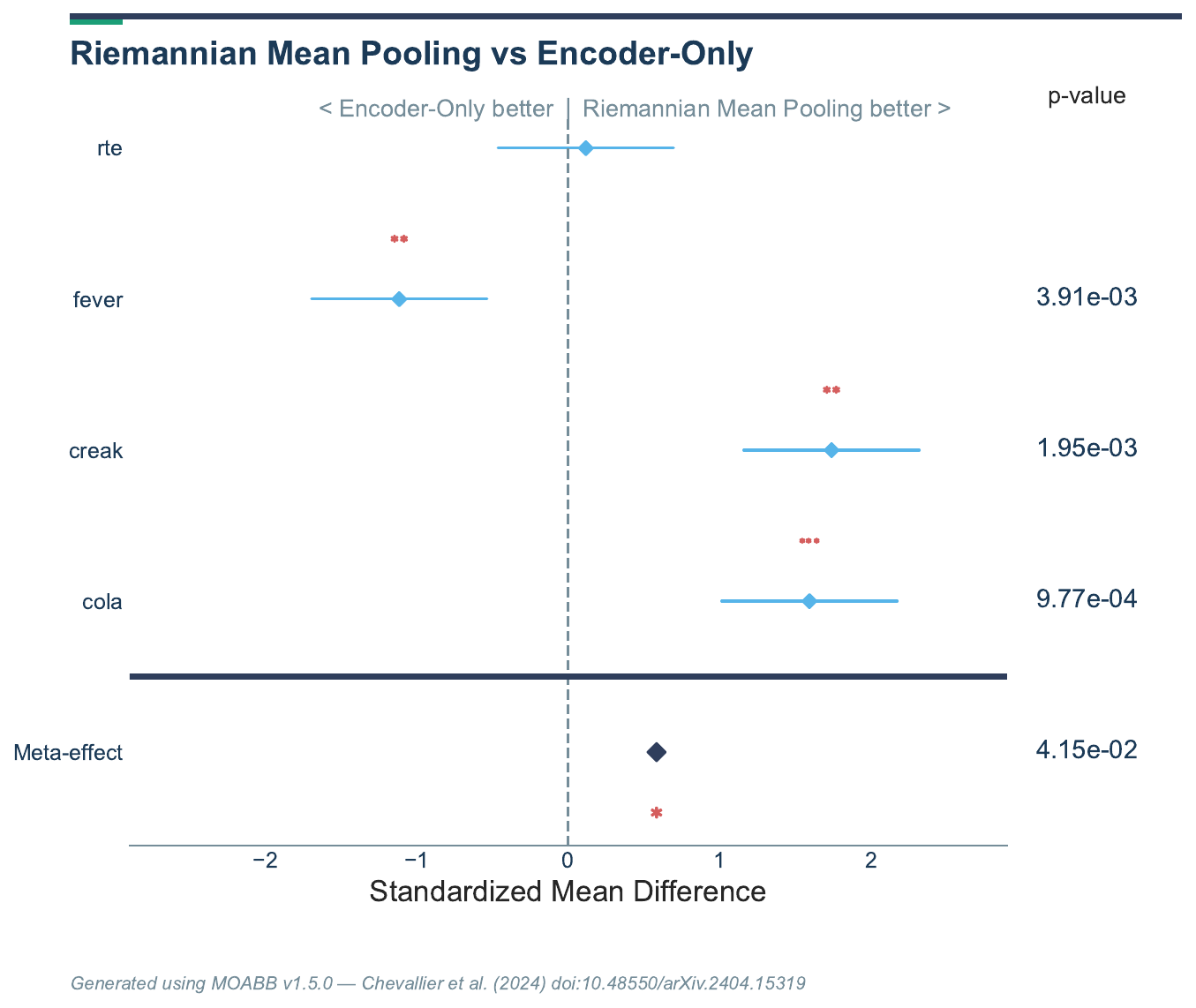}
  \caption{Riemannian Mean Pooling vs Encoder-Only: forest plot. Metric: \textbf{AUC}.}
  \label{fig:ablation-vs-encoder-only}
\end{figure}

\begin{figure}[H]
  \centering
  \includegraphics[width=\columnwidth]{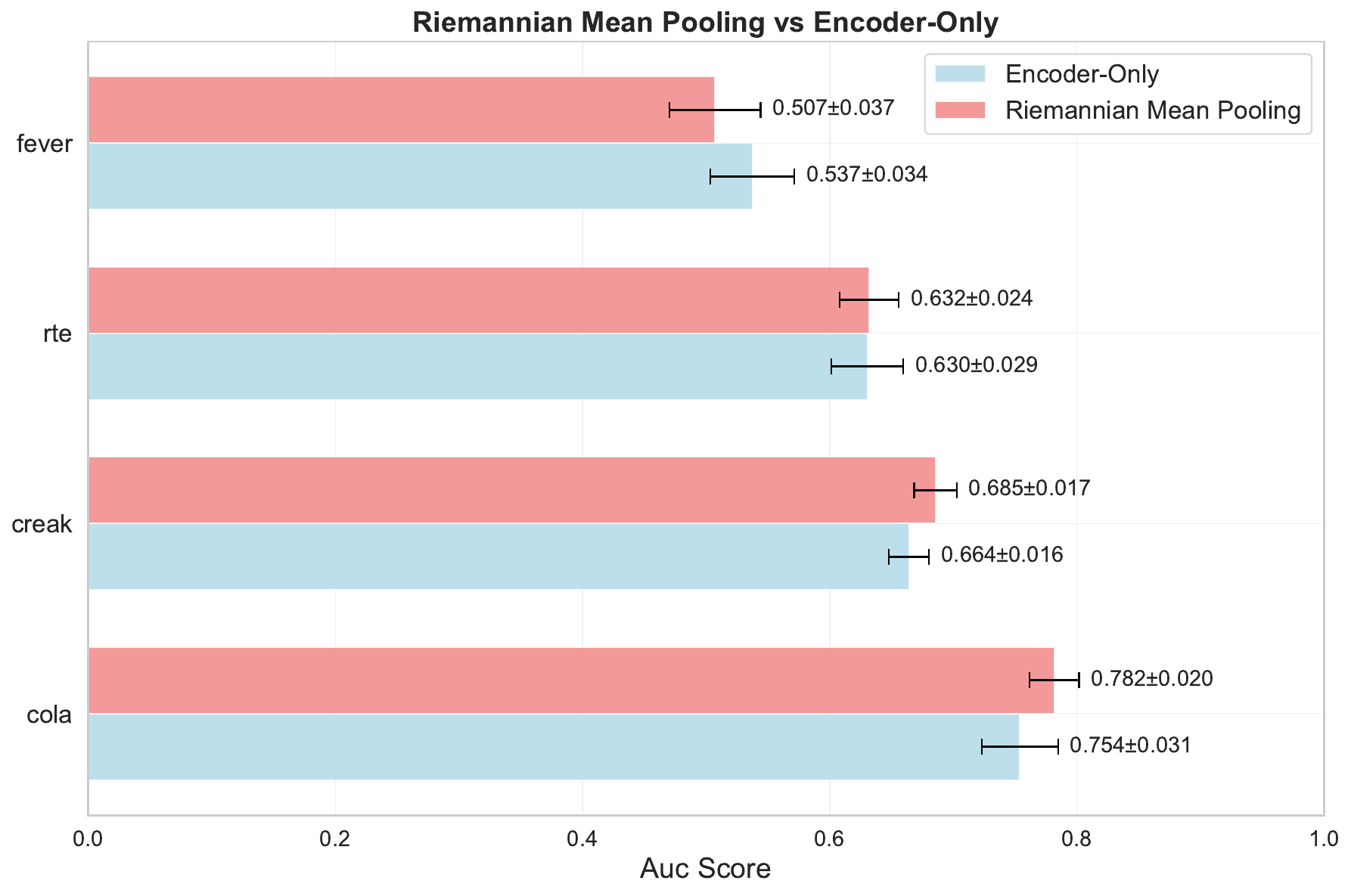}
  \caption{AUC bar comparison: Riemannian Mean Pooling vs Encoder-Only.}
  \label{fig:ablation-vs-encoder-only-barplot}
\end{figure}

\begin{figure}[H]
  \centering
  \includegraphics[width=\columnwidth]{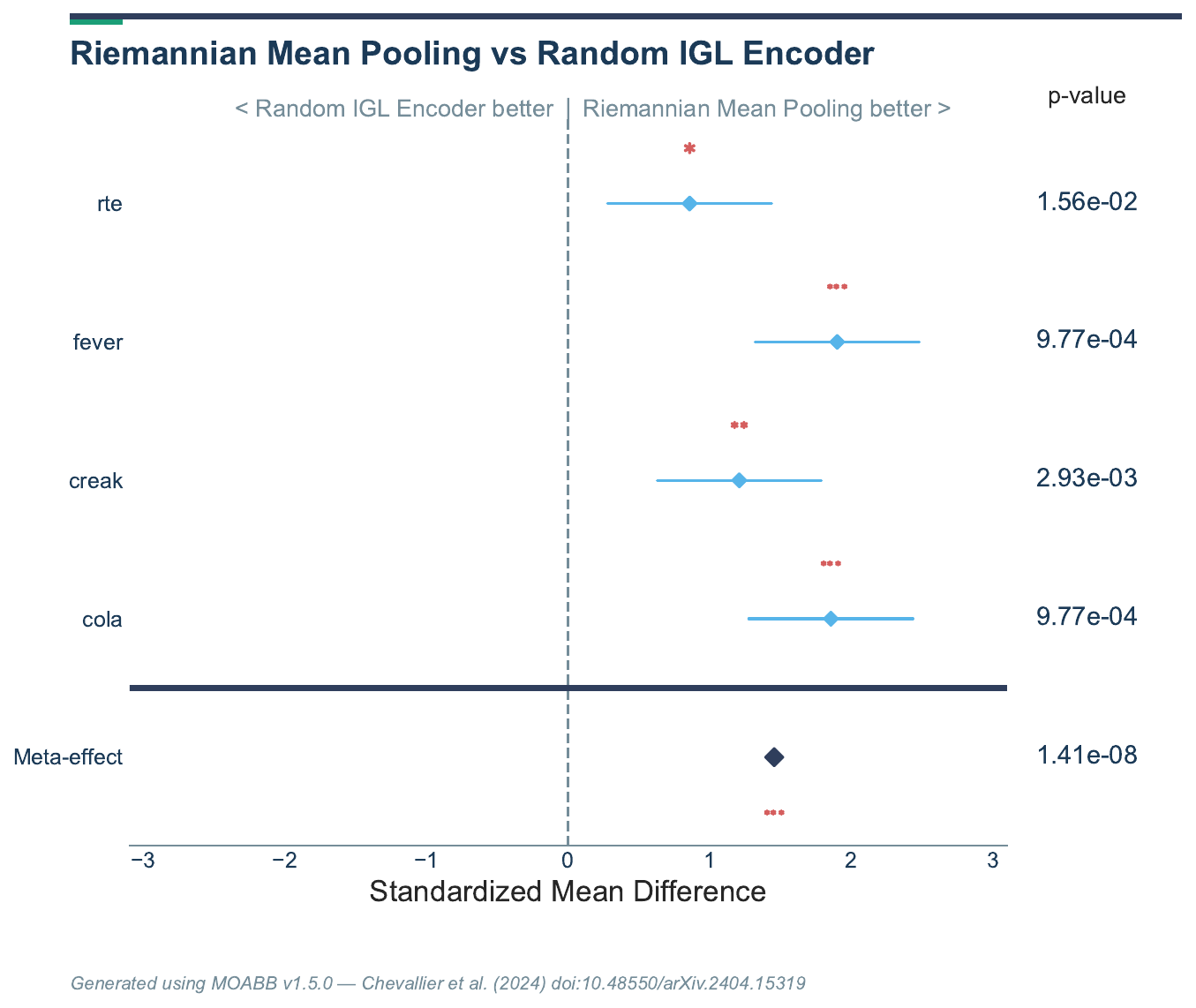}
  \caption{Riemannian Mean Pooling vs Frozen Random IGL: forest plot. Metric: \textbf{AUC}.}
  \label{fig:ablation-vs-frozen-random}
\end{figure}

\begin{figure}[H]
  \centering
  \includegraphics[width=\columnwidth]{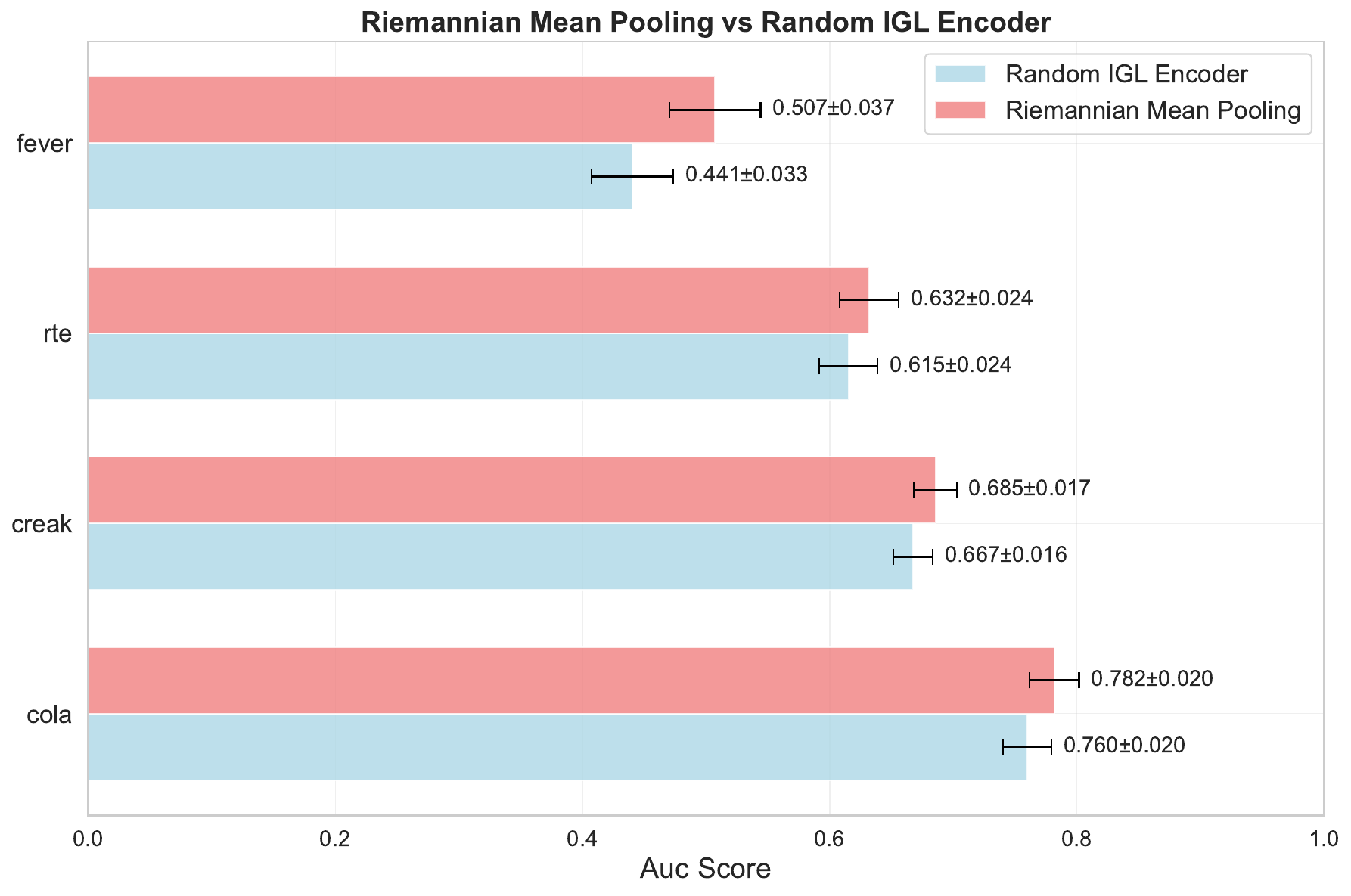}
  \caption{AUC bar comparison: Riemannian Mean Pooling vs Frozen Random IGL.}
  \label{fig:ablation-vs-frozen-random-barplot}
\end{figure}

\begin{figure}[H]
  \centering
  \includegraphics[width=\columnwidth]{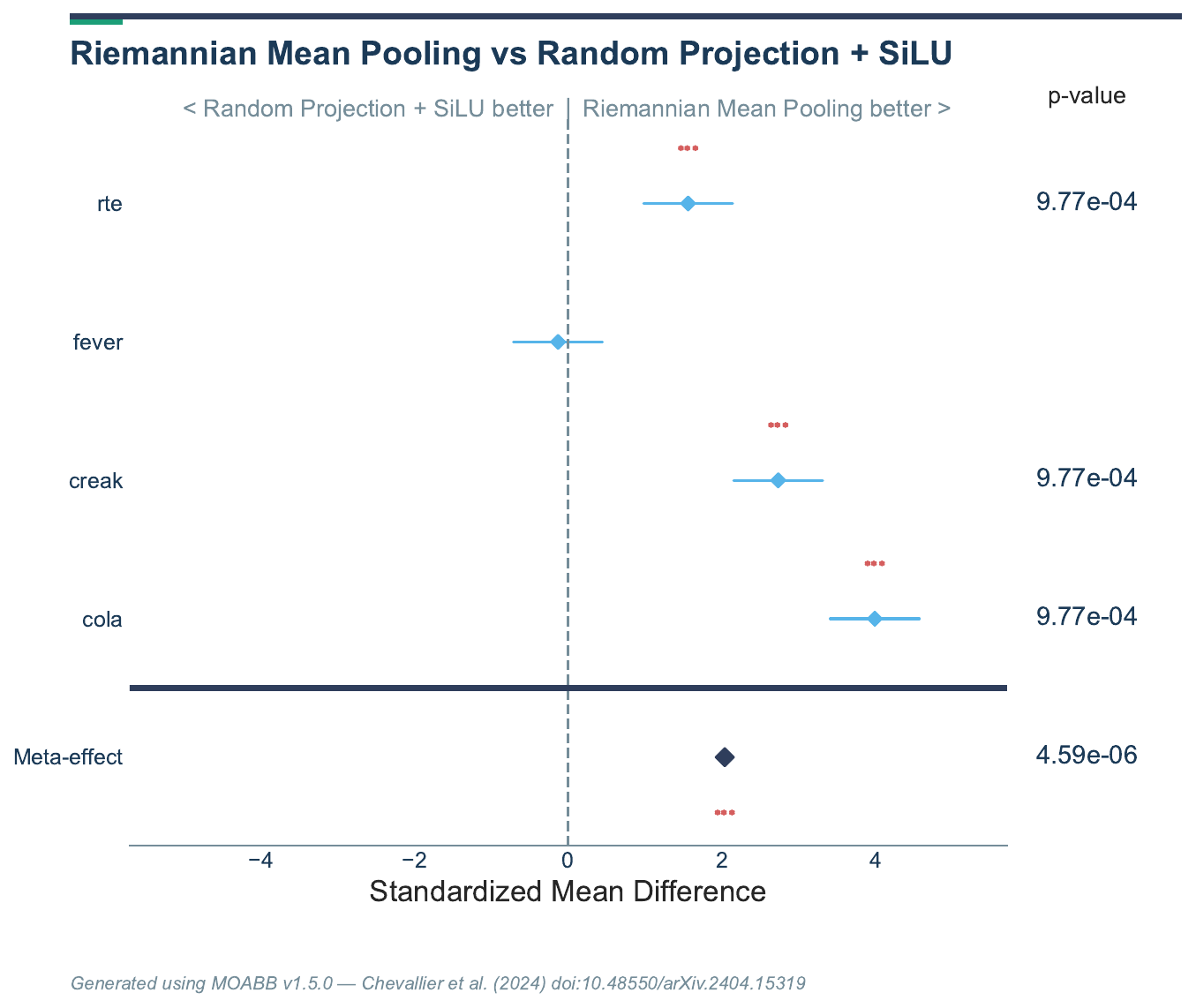}
  \caption{Riemannian Mean Pooling vs Random Projection + SiLU: forest plot. Metric: \textbf{AUC}.}
  \label{fig:ablation-vs-random-projection}
\end{figure}

\begin{figure}[H]
  \centering
  \includegraphics[width=\columnwidth]{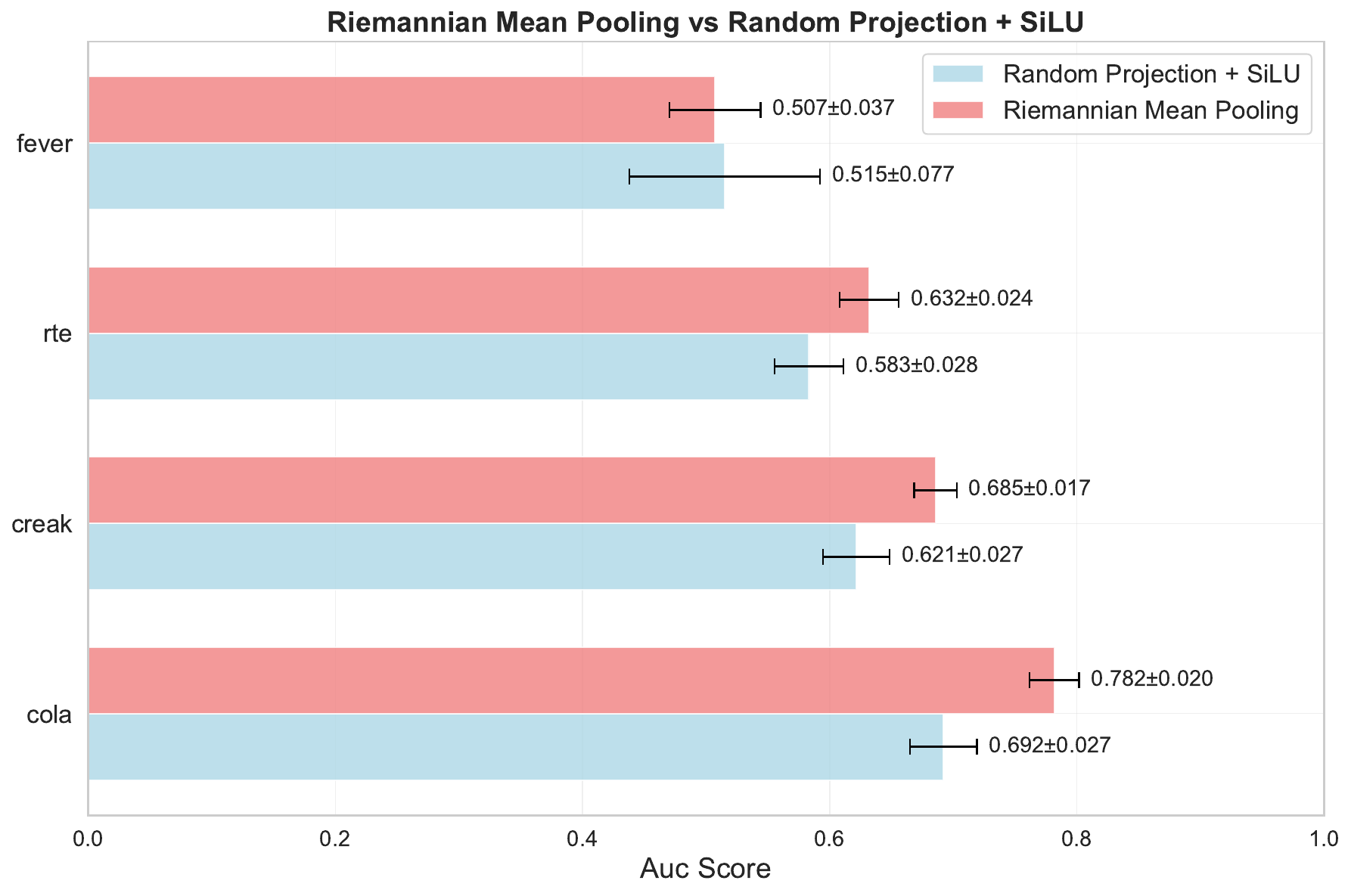}
  \caption{AUC bar comparison: Riemannian Mean Pooling vs Random Projection + SiLU.}
  \label{fig:ablation-vs-random-projection-barplot}
\end{figure}

\section{Additional results figures}
\label{app:additional-results}

The per-dataset AUC, Accuracy, and F1 bar plots together with the CLS-vs-RMP forest plot are deferred here. The body keeps the headline Linear-Probe-vs-RMP forest plot (Figure~\ref{fig:multiseed-statistical}) and Table~\ref{tab:multiseed-results} as the main visualisations.

\begin{figure}[H]
  \centering
  \includegraphics[width=\columnwidth]{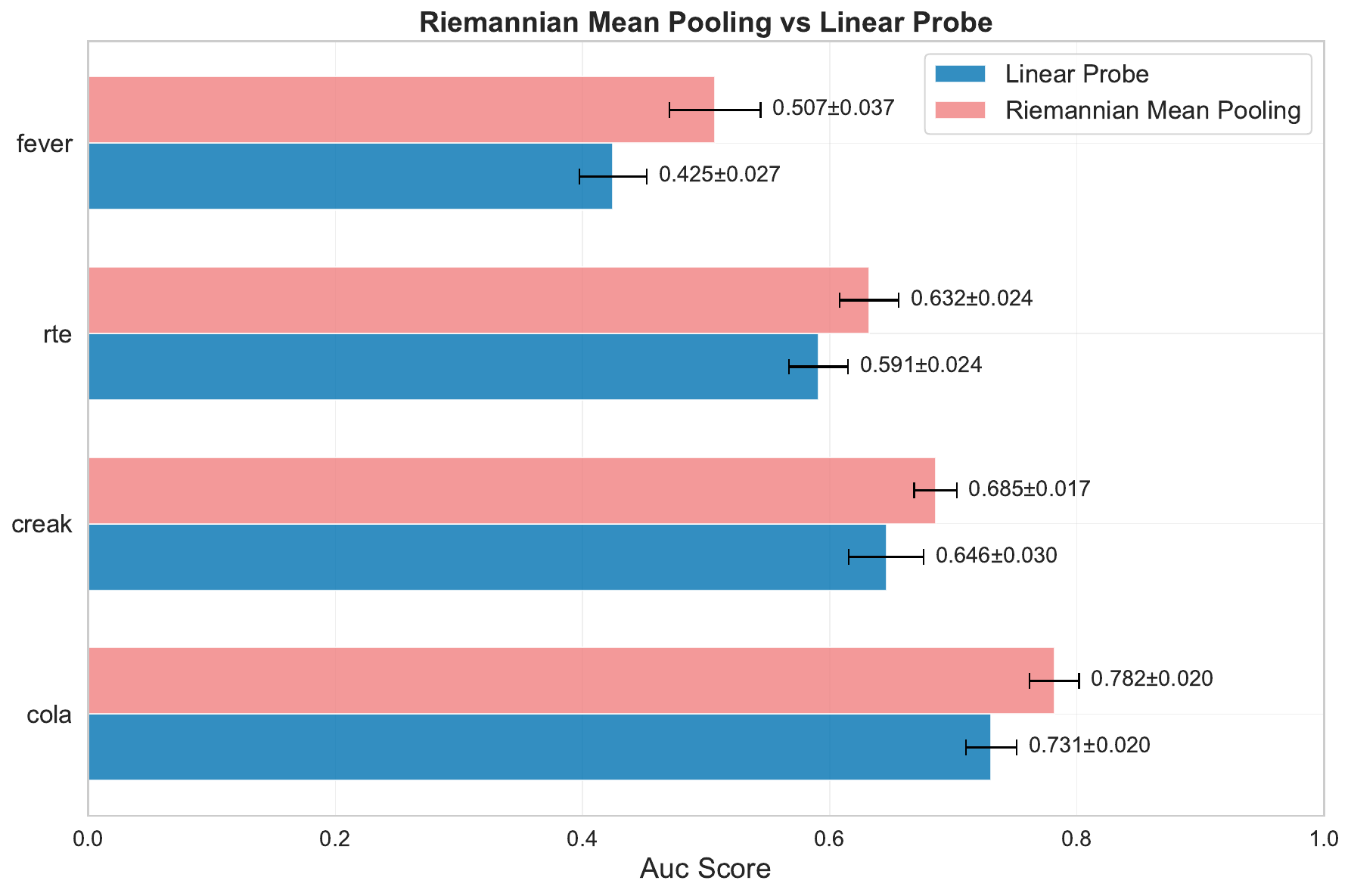}
  \caption{Multi-seed AUC comparison across datasets.}
  \label{fig:multiseed-auc}
\end{figure}

\begin{figure}[H]
  \centering
  \includegraphics[width=\columnwidth]{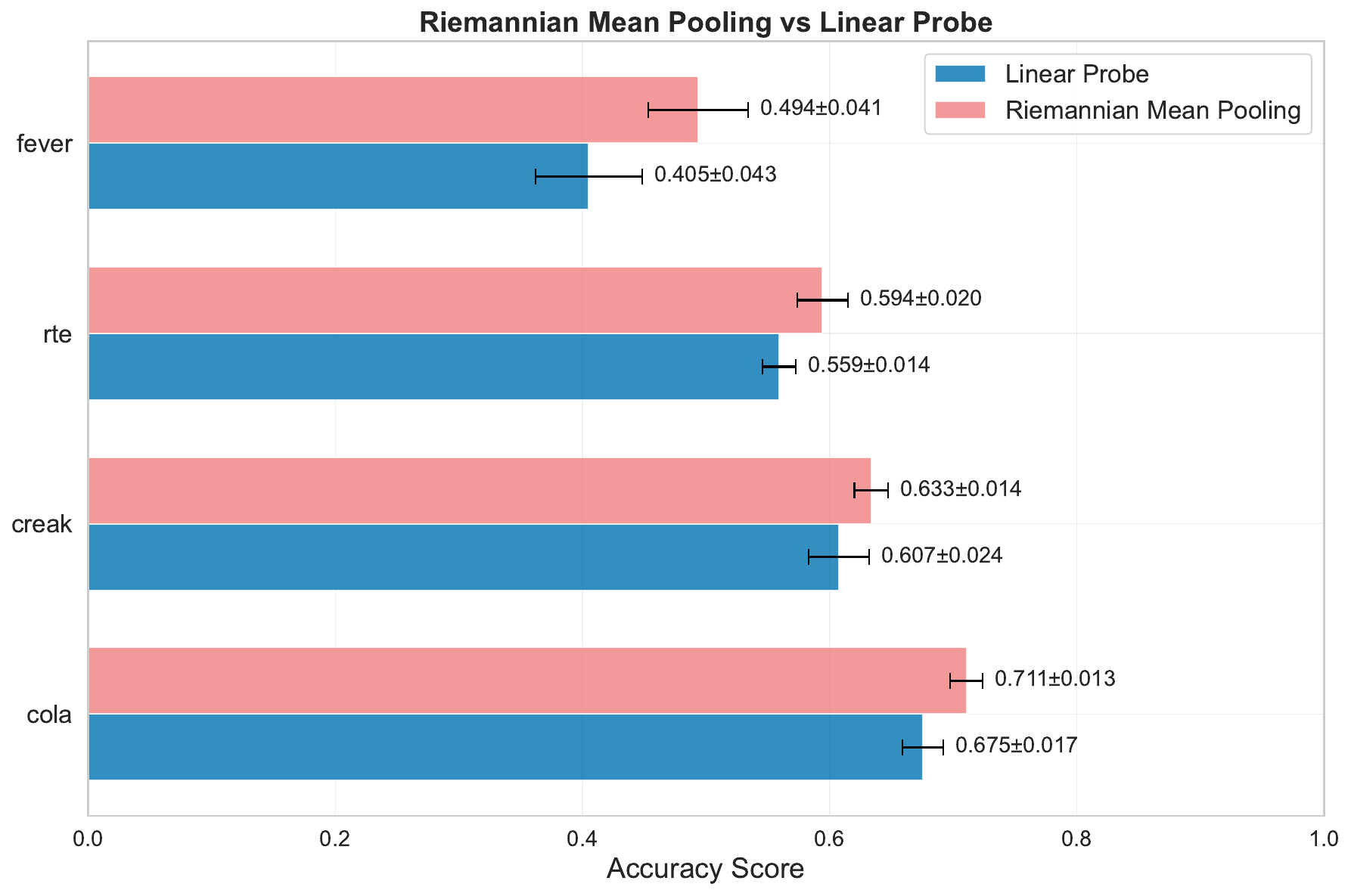}
  \caption{Multi-seed accuracy comparison across datasets. Error bars represent standard deviation across 10 evaluations.}
  \label{fig:multiseed-accuracy}
\end{figure}

\begin{figure}[H]
  \centering
  \includegraphics[width=\columnwidth]{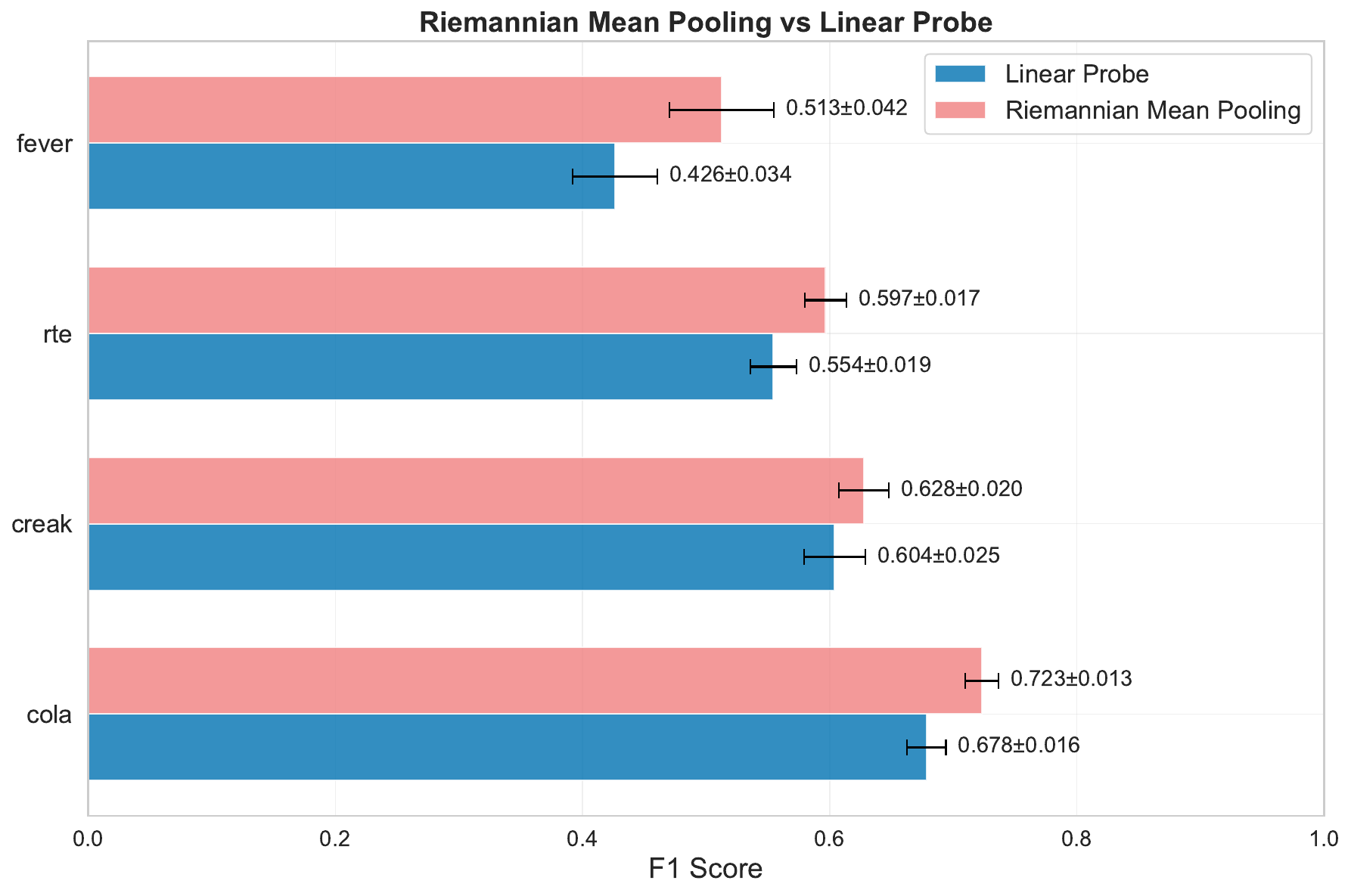}
  \caption{Multi-seed F1-score comparison across datasets.}
  \label{fig:multiseed-f1}
\end{figure}

\begin{figure}[H]
  \centering
  \includegraphics[width=\columnwidth]{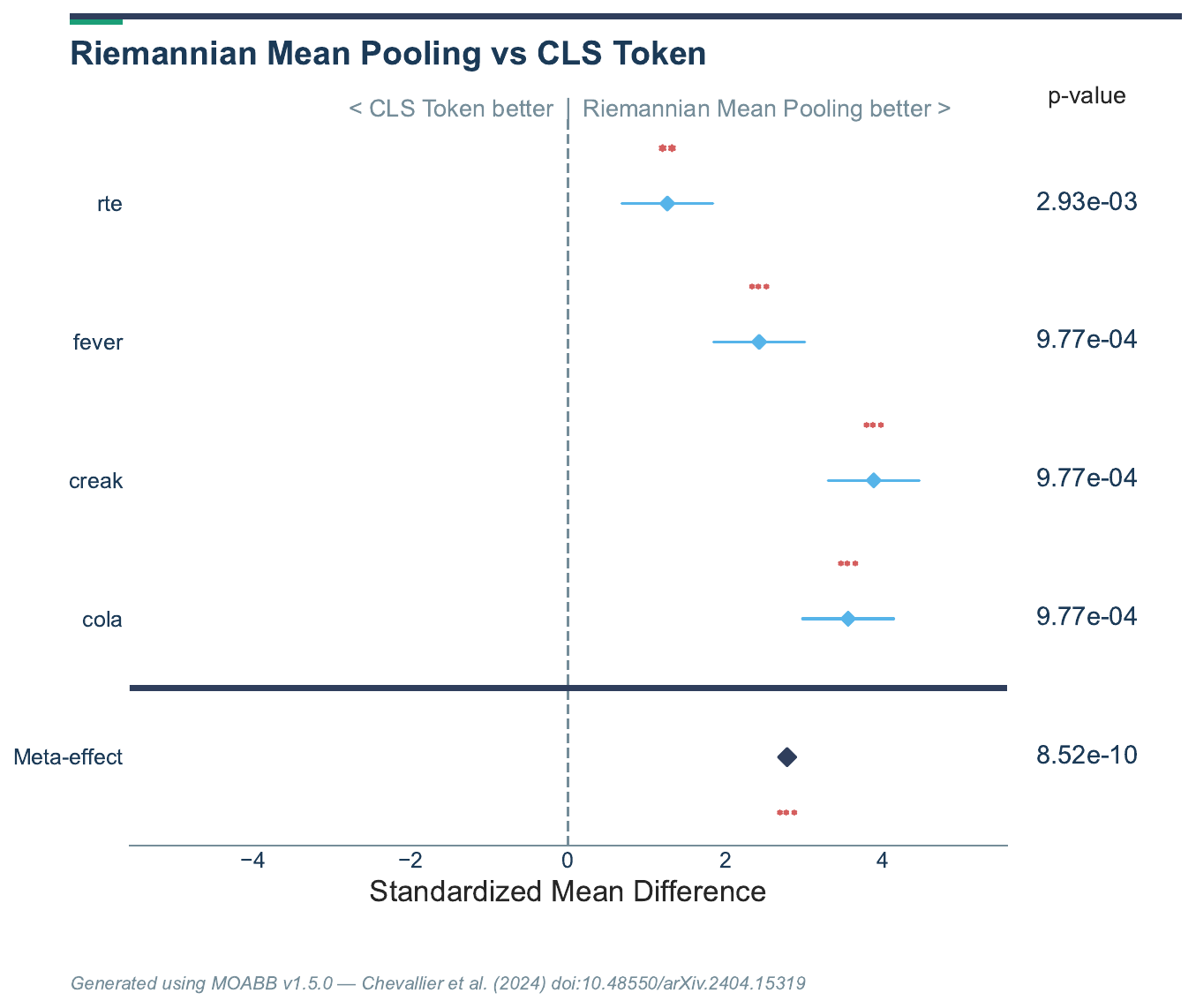}
  \caption{Statistical comparison between CLS Token Aggregation and RMP across random seeds. Box plots show the distribution of AUC scores from 10 evaluations (2 seeds $\times$ 5 folds). Effect sizes are reported using SMD. FEVER-Symmetric is included as a negative control (Section~\ref{sec:methodology}). A chance-level result is the expected and correct behaviour for both methods.}
  \label{fig:multiseed-statistical-cls}
\end{figure}

\section{The FEVER-Symmetric twin split}
\label{app:fever}

FEVER-Symmetric was built to remove the annotation artifacts of the original FEVER. Each claim is paired with a minimal edit of itself that carries the opposite label. The edit is sometimes a negation and more often a replaced noun or date, as in \emph{founded in 1971} against \emph{founded in 1991}, or \emph{the football championship of Africa} against \emph{the football championship of Asia}. Explicit negation words appear in only 22 of the 316 claims, so most pairs differ by a substitution rather than by a negation. A detector that relies on explicit negation alone has almost nothing to fire on and reaches $0.543$ AUC, close to chance, so the pairing leaves no lexical shortcut to the labels.

\citet{FEVERSymmetric} do not release the pair index, so we recover the pairing from claim text and treat it as an approximation. We obtain the pairings by matching each row to its opposite-label nearest neighbour at Jaccard overlap $\ge 0.6$, yielding 114 pairs (228 rows). The remaining 88 rows have no partner and no debiasing guarantee. The threshold is an arbitrary choice: 0.4, 0.5, and 0.7 yield 292, 282, and 168 paired rows respectively.

\begin{table}[H]
\centering
\small
\setlength{\tabcolsep}{3pt}
\begin{tabular}{lccc}
\toprule
Method & All & Twinned & Untwinned \\
 & (316) & (228) & (88) \\
\midrule
Linear Probe & 0.425 & 0.357 & 0.601 \\
CLS Token & 0.402 & 0.332 & 0.594 \\
Riemannian Mean Pooling & 0.509 & 0.445 & 0.662 \\
\bottomrule
\end{tabular}
\caption{FEVER-Symmetric split by whether a row has an opposite-label twin (AUC, mean across 10 evaluations). All three methods share the fold structure. Five-fold cross-validation separates a pair four times in five, so a method that relies on claim-surface features is trained on one member and evaluated on the other. The RMP rows come from the permutation run reported in this appendix rather than from the run behind Table~\ref{tab:multiseed-results}, so the full-set value reads $0.509$ here against $0.507$ there, a gap well inside the fold-to-fold standard deviation of $0.039$. The baseline rows are the same in both.}
\label{tab:fever-twin}
\end{table}

\paragraph{Inversion on the twinned rows}
Both Euclidean baselines fall far below chance on the rows that carry a twin, to $0.357$ and $0.332$, and rise above it on the rows that do not, to $0.601$ and $0.594$. This difference of $0.25$ AUC between two halves of the same dataset under the same model is consistent with the pairing construction, not with a property of the claims themselves.

\paragraph{Riemannian Mean Pooling on the twinned rows}
RMP also sits below chance on the twinned rows, at $0.445$, so it is not immune, yet it is significantly less affected. Paired over the 10 evaluations it exceeds Linear Probe by $0.088$ ($t = 4.63$, $p = 0.001$) and CLS Token Aggregation by $0.113$ ($t = 7.87$, $p = 2.5 \times 10^{-5}$) on those rows. On the untwinned rows the three methods lie within $0.07$ of each other and the differences are not significant. The full-set reading of $0.509$ falls between the two halves, weighted towards the 228 twinned rows against the 88 untwinned.

\paragraph{Residual signal in the Encoder-Only ablation}
The Encoder-Only ablation is the one method above chance on the full set, at $0.537 \pm 0.036$ ($p = 0.009$). It pools the trained encoder's coordinates in latent space rather than reading its Jacobian. The FEVER-Symmetric construction~\citep{FEVERSymmetric} targets surface features arising from the annotation process and does not target all model-accessible content, and BERT encodes substantial world knowledge through pre-training, so a Euclidean classifier over learned coordinates can detect a small residual signal. That signal is not specific to the geometric pipeline. In fact, on the untwinned rows the Linear Probe reaches $0.601$, so the content the debiasing leaves behind is accessible to mean pooling as well.

\section{Feature concatenation}
\label{app:concat}

A natural question is whether the geometric gain is a consequence of feature count. The tangent-space descriptor has 2080 coordinates against the 768 of a mean-pooled BERT vector, so a larger budget alone might account for it. Table~\ref{tab:concat} evaluates the two feature sets separately and concatenated, under one classifier. As in Table~\ref{tab:ablation-results}, the FEVER-Symmetric rows are reported for completeness and carry no information here, since no method on that dataset is expected to exceed chance. The comparisons below are over the three signal-bearing datasets.

\begin{table}[H]
\centering
\small
\setlength{\tabcolsep}{4pt}
\begin{tabular}{llc}
\toprule
Dataset & Features ($d$) & AUC \\
\midrule
\multirow{3}{*}{\textbf{FEVER}}
    & BERT only (768) & 0.425 $\pm$ 0.029 \\
    & Tangent space (2080) & 0.401 $\pm$ 0.059 \\
    & Both (2848) & 0.413 $\pm$ 0.035 \\
\midrule
\multirow{3}{*}{\textbf{CoLA}}
    & BERT only (768) & 0.731 $\pm$ 0.021 \\
    & Tangent space (2080) & 0.740 $\pm$ 0.013 \\
    & Both (2848) & 0.757 $\pm$ 0.020 \\
\midrule
\multirow{3}{*}{\textbf{CREAK}}
    & BERT only (768) & 0.646 $\pm$ 0.032 \\
    & Tangent space (2080) & 0.615 $\pm$ 0.031 \\
    & Both (2848) & 0.626 $\pm$ 0.035 \\
\midrule
\multirow{3}{*}{\textbf{RTE}}
    & BERT only (768) & 0.591 $\pm$ 0.025 \\
    & Tangent space (2080) & 0.592 $\pm$ 0.018 \\
    & Both (2848) & 0.599 $\pm$ 0.026 \\
\bottomrule
\end{tabular}
\caption{Feature concatenation on the whitened SPD descriptors (AUC, mean $\pm$ std across 10 evaluations). All three feature sets use the same standardised logistic regression, so the rows are comparable to each other. They are \emph{not} comparable to the Riemannian Mean Pooling rows of Table~\ref{tab:ablation-results}, which use pyRiemann's \texttt{TSClassifier} (tangent space followed by logistic regression, with no per-coordinate scaler). The BERT-only row is the Linear Probe of Table~\ref{tab:multiseed-results}, recomputed here as one half of the comparison, and reproduces it exactly.}
\label{tab:concat}
\end{table}

The largest feature set leads on CoLA and RTE while the smallest leads on CREAK, so the ordering is not monotone in the size of the feature set. With the classifier held fixed, more coordinates buy nothing consistent. Nearly three times as many gain $+0.009$ on CoLA ($p = 0.16$) and $+0.001$ on RTE ($p = 0.93$) while losing $0.031$ on CREAK ($p = 0.04$). In other words, AUC does not follow feature count.

Against the BERT-only features, concatenation gains $+0.027$ AUC on CoLA ($p = 0.0002$), loses $0.019$ on CREAK ($p = 0.03$), and does not move RTE ($+0.008$, $p = 0.13$). It is inconsistent in the same way, therefore we do not treat concatenation as a component of the method.

\section{Reconstruction convergence}
\label{app:recon-curve}

\begin{figure}[H]
  \centering
  \includegraphics[width=\columnwidth]{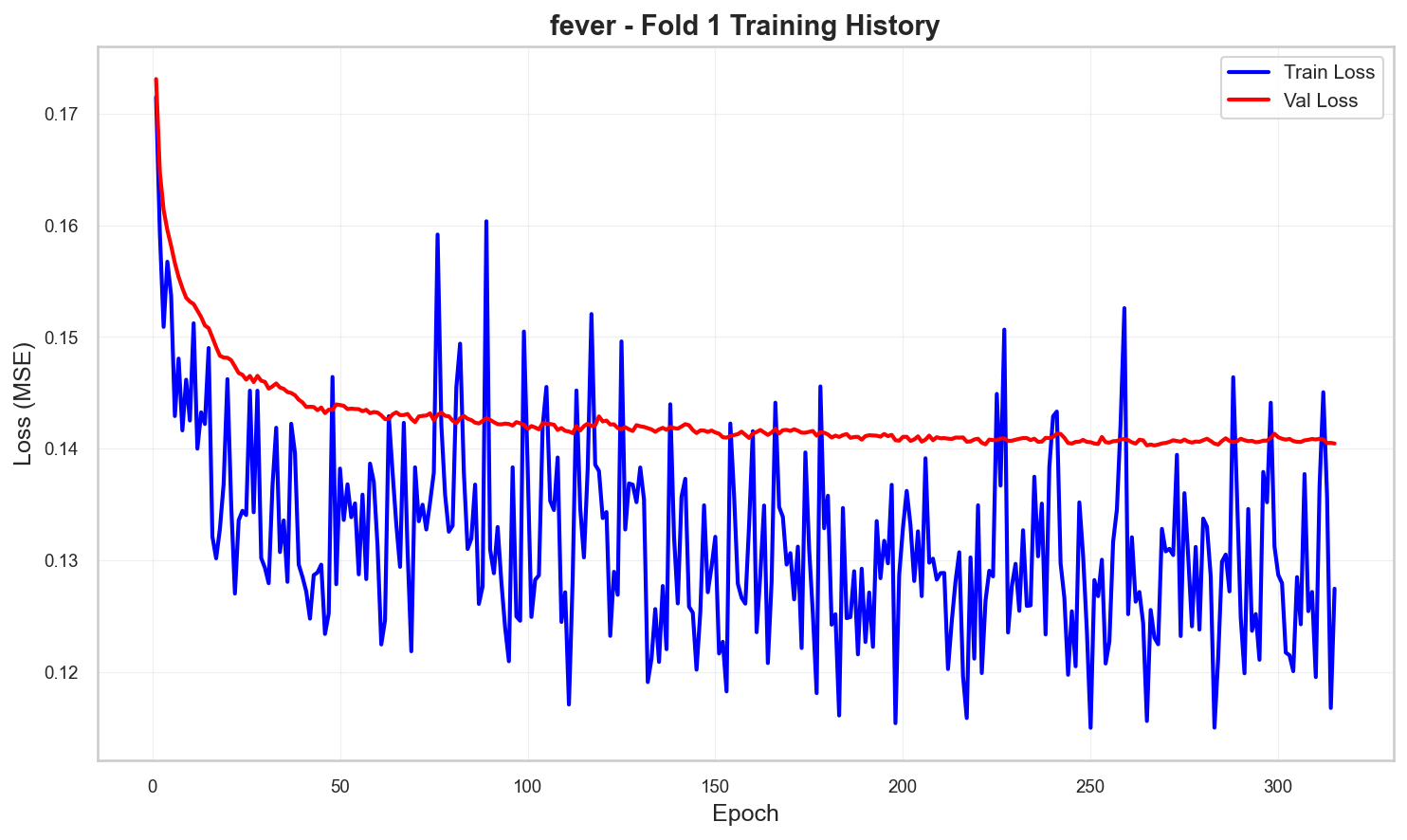}
  \caption{Training (blue) and validation (red) reconstruction MSE for the IGL encoder on fold 1 of FEVER-Symmetric, over $\sim$310 epochs. The validation curve decreases monotonically from $\sim$0.17 to $\sim$0.14. This is representative of all folds and datasets.}
  \label{fig:recon-curve}
\end{figure}

\begin{figure}[H]
  \centering
  \includegraphics[width=\columnwidth]{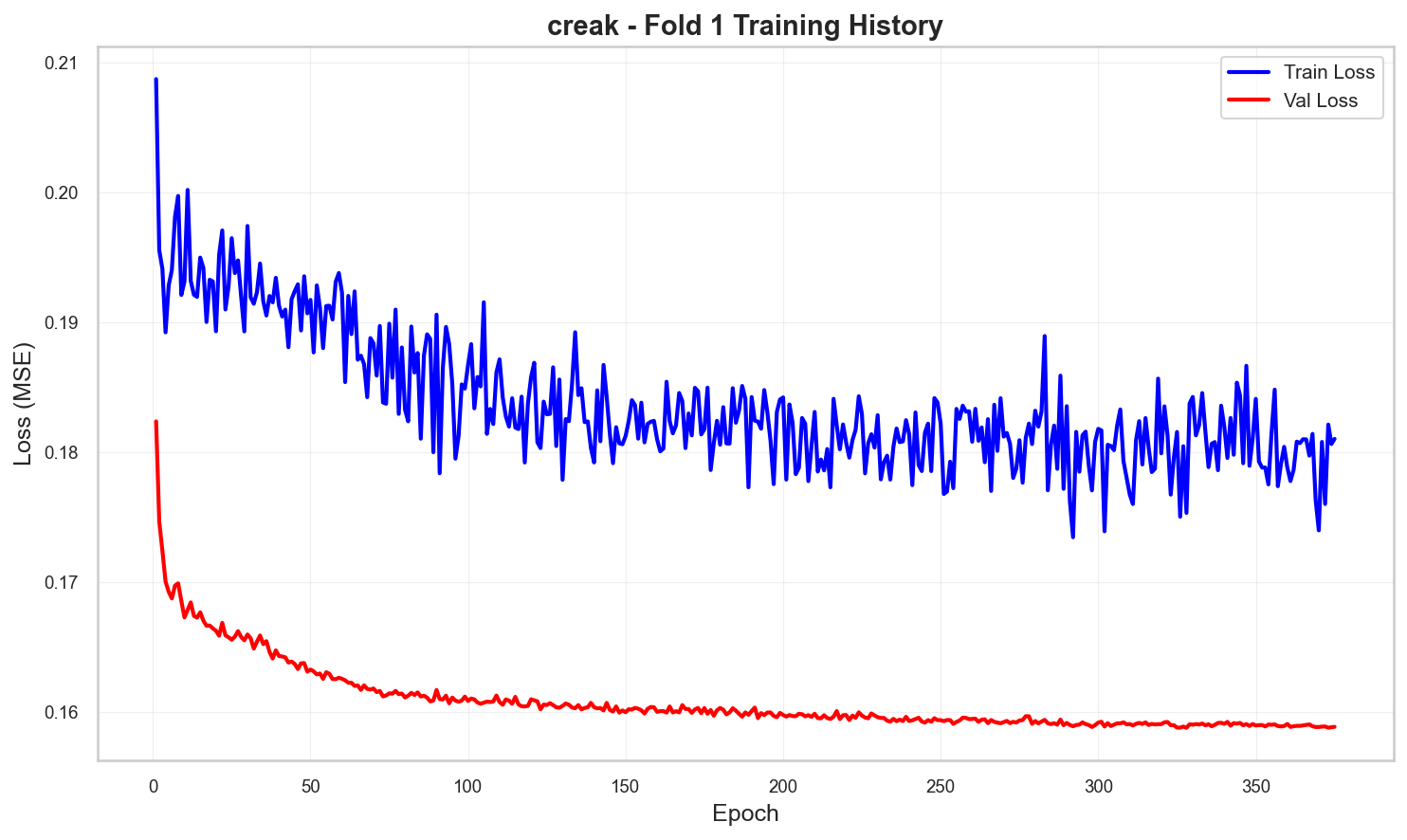}
  \caption{Training (blue) and validation (red) reconstruction MSE for the IGL encoder on fold 1 of CREAK, over $\sim$375 epochs. The validation curve decreases from $\sim$0.182 to $\sim$0.159.}
  \label{fig:recon-curve-creak}
\end{figure}

\section{Immersion}
\label{app:immersion}

Equation~\eqref{eq:cometric} treats $\Psi_\theta$ as a submersion. An alternative is to restrict it to the data manifold $M\subset\mathbb{R}^{768}$ of dimension $m < d_{\max}$, which makes it an immersion and to replace \eqref{eq:cometric} with 
\begin{equation}
g^{-1}(\mathbf{v}) = J(\mathbf{v}) \, P_M \, J(\mathbf{v})^\top,
\label{eq:immersion}
\end{equation}
where $P_M$ projects onto $T_{\mathbf{v}}M$. We estimate $P_M$ from the reconstruction readout as $J_R J_R^{+}$. This appendix reports what \eqref{eq:immersion} does to the descriptor. All runs are paired on fold, seed and encoder draw. They share one trained encoder per fold and differ only in the projector $P_M$.

\begin{figure*}[t]
  \centering
  \includegraphics[width=\textwidth]{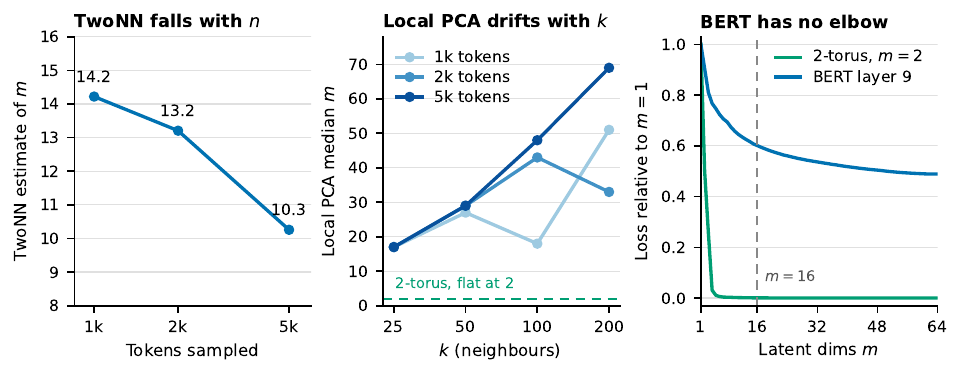}
  \caption{Stage 1 dimension estimates on CoLA layer-9 token embeddings (seed 67). \textbf{Left:} TwoNN falls as the sample grows, so it has not converged by $n = 5000$. \textbf{Centre:} the local PCA median rises with the neighbourhood size $k$ and does not agree across sample sizes, while the same estimator returns 2.0 at every $k$ on the torus control (dashed). \textbf{Right:} Matryoshka reconstruction loss against latent width, each curve divided by its own $m = 1$ value because the two absolute losses differ by two orders of magnitude. The torus curve reaches its floor by $m = 4$, while the BERT curve decays smoothly with no elbow to locate. We set $m = 16$ (dashed) for the variants in Table~\ref{tab:immersion}.}
  \label{fig:bert-dim}
\end{figure*}

\paragraph{Dimension estimates} No estimator we ran returned a stable dimension for layer-9 token embeddings, so $m$ is a choice rather than a measurement and the immersion variant is not identified by the data. TwoNN falls with sample size (14.2, 13.2, 10.3 at $n = 1000, 2000, 5000$). Maximum-likelihood estimates run 10.6 to 16.1 as $k$ goes from 5 to 50, while local PCA runs 17, 29, 48, 69 as $k$ goes from 25 to 200 at $n = 5000$ and does not agree across sample sizes. As a control we run the same estimators on a 2-torus embedded in $\mathbb{R}^{768}$, a surface of known dimension 2, where local PCA is flat at 2.0 for every $k$. The Matryoshka reconstruction curve, the estimator inherent to IGL, has no elbow. Figure~\ref{fig:bert-dim} plots TwoNN, local PCA and the Matryoshka curve.

\paragraph{Choice of $m$ and variants measured}
We set $m = 16$, which is a choice informed by the estimators rather than one derived from them. It is the top of the range where the only two estimators that agree with each other land, TwoNN at 10.3 to 14.2 across sample sizes and the maximum-likelihood estimate at 10.6 to 16.1 across $k$. The Matryoshka elbow detector returns 3 on a smooth monotone curve, which the gate rejects, and local PCA spans 17 to 69. We characterised the geometry at $m \in \{8, 16, 32, 48, 64\}$ and measured AUC at $m = 16$ alone, on CoLA (layer 9, 2 seeds $\times$ 5 folds, trained IGL encoder).
Table~\ref{tab:immersion} reports AUC, the numerical rank of $g^{-1}$ before regularisation, and the fraction of $J$'s Frobenius mass inside each run subspace (overlap). A sixth variant, \textsc{rank\_rot}, draws an independent rank-$m$ projector at each token. We omit it from the table because it collapses the Fr\'echet mean for reasons unrelated to the direction of the projection, which is the comparison \textsc{rank\_fixed} isolates by holding one projector fixed across tokens.

\begin{table}[H]
\centering
\small
\begin{tabular}{lrrr}
\toprule
Arm & AUC & Rank & Overlap \\
\midrule
\textsc{baseline} & 0.783 & 64 & 1.000 \\
\textsc{immersion} & 0.774 & 40 & 0.408 \\
\textsc{immersion\_trunc} & 0.776 & 16 & 0.089 \\
\textsc{rank\_fixed} & 0.784 & 16 & 0.023 \\
\textsc{immersion\_localpca} & 0.710 & 16 & 0.040 \\
\bottomrule
\end{tabular}
\caption{Immersion variants on CoLA ($m=16$, $n=10$ folds). The projector $P_M$ is the identity for \textsc{baseline}, $J_R J_R^{+}$ for \textsc{immersion}, the top-$m$ singular directions of $J_R$ for \textsc{immersion\_trunc}, a random shared rank-$m$ subspace for \textsc{rank\_fixed}, and an encoder-free local PCA estimate for \textsc{immersion\_localpca}. Rank is the numerical rank of \eqref{eq:immersion} before regularisation. Overlap is the fraction of $J$'s Frobenius mass inside the variant's subspace.}
\label{tab:immersion}
\end{table}

\paragraph{Comparison of the rank-16 variants}
The decisive comparison is \textsc{immersion\_trunc} against \textsc{rank\_fixed}. Both restrict to rank 16, but one uses the decoder's top-$m$ singular directions and the other an arbitrary fixed subspace.
\textsc{rank\_fixed} ties the baseline ($|\Delta\,\text{AUC}| < 0.001$, $p = 0.95$). \textsc{immersion\_trunc} sits below it ($\Delta\,\text{AUC} = -0.007$, $t = -2.38$, $p = 0.041$, uncorrected over the variant contrasts), so an arbitrary subspace of the same rank does as well as the decoder's.
The encoder-free tangent estimate (\textsc{immersion\_localpca}) is 0.074 below baseline, and its overlap with $J_R$'s column space is 0.040 against 0.023 for the random rank-16 control, so the two estimates of $T_\mathbf{v}M$ are near-orthogonal.

\paragraph{Rank of the immersion variant}
The rank column of Table~\ref{tab:immersion} should be read with the same caution as $m$. $J_R$ has no spectral gap. Its singular values decay smoothly from 26.4 to 2.7 (a condition number of 9.6), and the largest ratio between consecutive values anywhere in the spectrum is 1.2. Counting the directions that carry 90\% of the energy gives 38, counting the values above numerical zero gives 64, and a second run of the same variant on the same dataset reports 64 where the run in Table~\ref{tab:immersion} reports 40. On the torus control the same estimator leaves no such ambiguity, with 3 directions carrying 90\% of the energy and a factor of $1.1 \times 10^{4}$ between the fifth and sixth singular values. The decoder's estimate of the tangent space is therefore no better identified than $m$ is, which is consistent with the projector making no difference to AUC.

\paragraph{Numerical conditioning}
In \texttt{float32} the metrics lose positive-definiteness on rank-deficient matrices of this kind. Eigenvalues reach $-2.2\times 10^{-7}$ against machine epsilon $1.2\times 10^{-7}$. In such a scenario eigenvalues are numerically zero within arithmetic precision. While the regularised matrix is practically invertible, the regulariser dominates, and all variants tie for numerical reasons. We therefore perform the inversion in \texttt{float64}, where the regularisation fire rate is 0.0\% across every run and every $m$. The lack of improvement reflects geometry rather than numerical breakdown.

Restricting the metric to an estimated manifold therefore does not help, and since neither $m$ nor $T_{\mathbf{v}}M$ is identified on these embeddings, we report the submersion metric of \eqref{eq:cometric} throughout.

\end{document}